\documentclass[lettersize,journal]{IEEEtran}
\usepackage{amsmath,amsfonts}
\usepackage{algorithmic}
\usepackage{algorithm}
\usepackage{array}
\usepackage[caption=false,font=normalsize,labelfont=sf,textfont=sf]{subfig}
\usepackage{textcomp}
\usepackage{stfloats}
\usepackage{url}
\usepackage{verbatim}
\usepackage{graphicx}
\usepackage{cite}
\usepackage{threeparttable}
\usepackage{makecell}
\usepackage{wrapfig}
\usepackage{float}
\usepackage{multirow}
\usepackage{footnote}
\usepackage{booktabs}
\begin{document}

\title{Which Samples Should be Learned First: Easy or Hard?}

\author{Xiaoling Zhou and Ou Wu
\thanks{
Xiaoling Zhou and Ou Wu are with the Center for Mathematics, Tianjin University, Tianjin, China. Ou Wu is the corresponding anthor. Email addresses: xiaolingzhou@tju.edu.cn (Xiaoling Zhou), wuou@tju.edu.cn (Ou Wu)}
\thanks{Manuscript received x xx, xxxx; revised x xx, xxxx.}}

\markboth{Journal of \LaTeX\ Class Files,~Vol.~xx, No.~x, x~xxxx}%
{Shell \MakeLowercase{\textit{et al.}}: A Sample Article Using IEEEtran.cls for IEEE Journals}


\maketitle

\begin{abstract}
An effective weighting scheme for training samples is essential for learning tasks. Numerous weighting schemes have been proposed. Some schemes take the easy-first mode, whereas some others take the hard-first one. Naturally, an interesting yet realistic question is raised. Which samples should be learned first given a new learning task, easy or hard? To answer this question, both theoretical analyses and experimental verification are conducted. First, a general optimized objective function is proposed, revealing the relationship between the difficulty distribution and the difficulty-based sample weights. Second, on the basis of the optimized objective function, theoretical answers are obtained. Besides the easy-first and hard-first modes, there are two other priority modes, namely, medium-first and two-ends-first. The prior mode does not necessarily remain unchanged during the training process. Third, an effective and universal solution is proposed to select the optimal priority mode when there is no prior knowledge or theoretical clues. The four modes, namely, easy/medium/hard/two-ends-first, can be flexibly switched in the proposed solution. Fourth, a wide range of experiments is conducted under various scenarios to further compare the weighting schemes in different modes. On the basis of these works, reasonable and comprehensive answers are obtained. Factors including the distribution of samples' learning difficulties and the validation data determine which samples should be learned first in a learning task. 
\end{abstract}

\begin{IEEEkeywords}
priority mode, learning difficulty, weighting scheme, easy-first, hard-first, bias-variance trade-off.
\end{IEEEkeywords}

\section{Introduction}
\IEEEPARstart{I}{t} is widely accepted that model training is sensitive to the weights of training samples. Treating each sample unequally can improve the learning performance. The cues and inspirations for the design of the weighting function in a weighting scheme are usually derived from the following two aspects:
\begin{itemize}
\item Application contexts of the learning task. Tasks such as fraud detection~\cite{fraud001} and medical diagnosis~\cite{medical} are cost-sensitive. Different samples have unequal importance according to their gains or costs. Therefore, samples with high gains/costs will be assigned high weights.
\item Characteristics of the training data. Training samples are different from each other in characteristics, such as data quality
~\cite{BuyuLi11, Verleysen91}, sample neighbors~\cite{JingfengZhang16, DeliChen35}, and category distribution
~\cite{YinCui17, detectionsurvey, Fernando_imb}. In some tasks, some labels are of low confidence or with high noises, so these samples should be assigned low weights. In some other tasks, samples in the minority categories are usually more difficult to learn well, so these samples should be assigned high weights.
\end{itemize}

Context-inspired weighting functions are usually defined in a heuristic manner and only used in special applications, whereas characteristics-inspired weighting functions have received increasing attention in recent years due to their effectiveness and universality. Data characteristics are related to an important property of samples, namely, learning difficulty. Most related studies split training samples into easy/hard or easy/medium/hard according to samples' learning difficulties. In some schemes, hard samples are assigned high weights, which is called the hard-first mode. For example, Lin et al.~\cite{TsungYiLin03} proposed Focal loss in object detection, which significantly improves the detection performance. In some other schemes, easy samples have higher weights than hard ones, which is called the easy-first mode. Kumar et al.~\cite{MPawanKumar04} proposed self-paced learning (SPL), which sets the weights of hard samples to zero with a threshold. The threshold is gradually increased to ensure that more hard samples can participate in the training. These two priority modes, namely, easy-first and hard-first, appear to contradict each other yet both demonstrate effectiveness in certain learning tasks. Consequently, a natural question is raised. Which samples should be learned first facing a new task, easy or hard ones? Indeed, several studies have proposed similar concerns. For example, Wang et al.~\cite{XinWang22} raised a similar question about ``easy-first versus hard-first" under the context of curriculum learning. 

To answer the question (called the ``easy-or-hard" question), partial observations and conclusions in existing studies are summarized and preliminary answers are obtained. Both theoretical analysis and empirical verification are then performed. More comprehensive answers are presented. Our contributions are summarized as follows:
\begin{itemize}
    \item Preliminary observations and conclusions for the investigated question from existing studies are summarized. Four unsolved subproblems are raised according to the gap between the question and the summarized preliminary answers.
    \item To theoretically explore the ``easy-or-hard" question, a general optimized objective function is constructed for difficulty-based sample weights. It reveals the relationship between the learning difficulty distribution and the priority mode. Theoretical analysis based on the bias-variance trade-off theory is then carried out and some answers are obtained for existing concerns.
    \item An effective data-adaptive weighting solution is proposed for selecting the optimal priority mode when there is no prior knowledge or theoretical clues.  
Compared with existing methods, our proposed solution can be flexibly switched among four priority modes, namely, easy-first, medium-first, hard-first and two-ends-first. In contrast, existing weighting methods can achieve only one of the four modes.
\item Extensive experiments on 
various tasks under different occasions (i.e., noisy, imbalanced, standard, and different difficulty distribution) are conducted on benchmark data sets. The empirical observations further support our main theoretical conclusions. In addition, our proposed weighting function achieves competitive results in all the above typical learning scenarios.
\end{itemize}

\section{Related studies}
\subsection{Notations}
We define the symbols including the main symbols in Table~1 as follows. Let $T = \{(x_{i}, y_{i})\}_{i=1}^N$ be a set of $N$ training samples, where $x_{i}$ is the input feature and $y_{i}$ is the associated label. Let $C$ be the number of categories and $y_{i} \in \{1,\dots, C\}$. Let $r_{c}$ be the empirical class frequency of the $c$-th category. Let $\mathcal{L}$ be the training loss. $w_{i}$ and $l_{i}$ are the weight and loss of the $i$-th sample. $d_{i}$ is the $i$-th sample's learning difficulty. Let $p \in [0,1]$ be the predicted probability for the correct category. $w^{c}$ is the weight of the $c$-th category when the category-wise weighting strategy is used.
\subsection{Existing weighting schemes}
\begin{table*}[t]
\caption{Several typical weighting schemes.}
\label{tab1}
\centering
\resizebox{1.9\columnwidth}{!}{
\begin{tabular}{|c|c|c|c|c|c|c|}
\hline
Method                                                           & Weighting scheme & Domain                                                & Scenario                                                                                                                     & Measurement                                                             & Priority mode                                                                                 & Granularity \\ \hline
\begin{tabular}[c]{@{}c@{}}SPL\_ \\ Binary~\cite{MPawanKumar04} \end{tabular}          & $\min _{w \in[0,1]^{n}} \mathcal{L}(w, \lambda, l)=\sum_{i=1}^{n} w_{i} l_{i}-\lambda \sum_{i=1}^{n} w_{i}$                & \begin{tabular}[c]{@{}c@{}}NLP\\ CV\end{tabular}      & \begin{tabular}[c]{@{}c@{}}Noun Phrase Coreference \\ Image classification \\ Object Localization \\ (Standard)\end{tabular} & Loss                                                                  & Easy-first                                                                                    & Sample      \\ \hline
 SPL\_Log~\cite{LuJiang06}                                                     & $\min _{w \in[0,1]^{n}} \mathcal{L}(w, \lambda, l)=\sum_{i=1}^{n} w_{i} l_{i}+\sum_{i=1}^{n}\left(\xi w_{i}-\xi^{w_{i}} / \log \xi\right), \xi=1-\lambda$                & CV                                                    & \begin{tabular}[c]{@{}c@{}}Multimedia Event Detection \\ (Standard)\end{tabular}                                             & Loss                                                                  & Easy-first                                                                                    & Sample      \\ \hline
\begin{tabular}[c]{@{}c@{}}Cost-\\ sensitive \\ SPL~\cite{MaciejZieba07}     \end{tabular} & $w_{i}=\left\{\begin{array}{l}
1, \textit { if } l_{i}<y_{i} C_{+}+\left(1-y_{i}\right) C_{-} \\
0, \textit { otherwise }
\end{array}\right.$                & CV                                                    & \begin{tabular}[c]{@{}c@{}}Image classification \\ (Imbalanced)\end{tabular}                                                  & Loss                                                                  & Easy-first                                                                                    & Mixture     \\ \hline
Focal loss~\cite{TsungYiLin03}                                                                                                          &  $\mathcal{L}(\gamma)=-(1-p)^{\gamma} \log (p)  $             & CV                                                    & \begin{tabular}[c]{@{}c@{}}Dense Object Detection \\ (Imbalanced)\end{tabular}                                                & Pred                                                                  & Hard-first                                                                                    & Sample      \\ \hline
QFL~\cite{XiangLi09}                                                                                                                 & $\mathcal{L}(\sigma, \beta)=\sum_{i=1}^{N}\left(-\left|y_{i}-\sigma\right|^{\beta}\left(\left(1-y_{i}\right) \log (1-\sigma)+y_{i} \log (\sigma)\right)\right) $               & CV                                                    & \begin{tabular}[c]{@{}c@{}}Dense Object Detection \\ (Imbalanced)\end{tabular}                                                & Pred                                                                  & Hard-first                                                                                    & Sample      \\ \hline
 ASL~\cite{EmanuelBenBaruch10}                                                            & $l_{i}\left(\gamma_{+}, \gamma_{-}, m\right)=\left\{\begin{array}{l}
\left(1-p_{i}\right)^{\gamma_{+}} \log \left(p_{i}\right), y_{i}=1 \vspace{2pt}\\
\left(p_{i, m}\right)^{\gamma_{-}} \log \left(1-p_{i, m}\right), y_{i}=0
\end{array}, p_{i, m}=\max \left(p_{i}-m, 0\right)\right.$                & CV                                                    & \begin{tabular}[c]{@{}c@{}}Dense Object Detection \\ (Imbalanced)\end{tabular}                                                & Pred                                                                  & \begin{tabular}[c]{@{}c@{}}Hard-first
\end{tabular} & Sample      \\ \hline
 AdaBoost~\cite{YoavFreund12}                                                       & $w_{i}^{m+1}=w_{i}^{m} \exp \left(\alpha_{m}\right) $               & CV                                                    & \begin{tabular}[c]{@{}c@{}}Handwritten Digit Recognition \\ (Standard)\end{tabular}                                          & Error (Loss)                                                                 & Hard-first                                                                                    & Sample      \\ \hline
G-RW~\cite{SongyangZhang14}                                                                                  & $w^{c}=\left(1 / r_{c}\right)^{\rho} / \sum_{k=1}^{C}\left(1 / r_{k}\right)^{\rho}  $              & CV                                                    & \begin{tabular}[c]{@{}c@{}}Image classification \\ Object detection \\ (Imbalanced)\end{tabular}                              & \begin{tabular}[c]{@{}c@{}}Empirical\\ class\\ frequency\end{tabular} & Hard-first                                                                                    & Category    \\ \hline
GAIRAT~\cite{JingfengZhang16}                                                              & $w_{i}=\left(1+\tanh \left(\lambda+5 \times\left(1-2 \times k\left(x_{i}, y_{i}\right) / K\right)\right)\right) / 2  $              & CV                                                    & \begin{tabular}[c]{@{}c@{}}Image classification \\ (Standard)\end{tabular}                                                   & \begin{tabular}[c]{@{}c@{}}Margin\end{tabular}  & Hard-first                                                                                    & Sample      \\ \hline
 \begin{tabular}[c]{@{}c@{}}Class-\\ balance~\cite{YinCui17}           \end{tabular}         & $w^{c}=(1-\beta) /\left(1-\beta^{N_{c}}\right), \beta \in[0,1)$                & CV                                                    & \begin{tabular}[c]{@{}c@{}}Image classification \\ (Imbalanced)\end{tabular}                                                  & \begin{tabular}[c]{@{}c@{}}Category\\ Proportion\end{tabular}         & Hard-first                                                                                    & Category    \\ \hline
\begin{tabular}[c]{@{}c@{}}Truncated \\ Loss~\cite{WenjieWang18}          \end{tabular}        & $l_{i}=\left\{\begin{array}{l}
0,~l_{i}^{C E}>0 \wedge  y_{i}=1 \vspace{2.5pt}\\
l_{i}^{C E}, \textit { otherwise }
\end{array}\right.$                & \begin{tabular}[c]{@{}c@{}}Data\\ mining\end{tabular} & \begin{tabular}[c]{@{}c@{}}Recommendation \\ (Noisy)\end{tabular}                                                            & Loss                                                                  & \begin{tabular}[c]{@{}c@{}}Easy-first
\end{tabular}          & Mixture     \\ \hline
 LOW~\cite{CarlosSantiagoa19}                                                             & $R(w ; \lambda)=-w^{T} \nabla_{\theta_{t}}+\lambda\|w-1\|^{2}$                & CV                                                    & \begin{tabular}[c]{@{}c@{}}Image classification \\ (Imbalanced)\end{tabular}                                                  & Gradient                                                              & Hard-first                                                                                    & Sample      \\ \hline
JTT~\cite{EvanZheranLiu20}                                                                                                                & $\mathcal{L}(l, E)=\left(\lambda_{u p} \sum_{\left(x_{i}, y_{j}\right) \in E} l_{i}+\sum_{\left(x_{j}, y_{j}\right) \notin E} l_{j}\right) $               & \begin{tabular}[c]{@{}c@{}}NLP\\ CV\end{tabular}      & \begin{tabular}[c]{@{}c@{}}Image classification \\ Sentiment analysis \\ (Standard)\end{tabular}                             & Loss                                                                  & Hard-first                                                                                    & Partial data \\ \hline
SuperLoss~\cite{ThibaultCastells21}                                                           & $\mathcal{L}\left(l_{i}, \sigma_{i}\right)=\left(l_{i}-\tau\right) w_{i}+\lambda\left(\log w_{i}\right)^{2}$                & CV                                                    & \begin{tabular}[c]{@{}c@{}}Object detection\\Image retrieval \\ (Noisy)\end{tabular}                                         & Loss                                                                  & Easy-first                                                                                    & Sample      \\ \hline
DWB~\cite{Fernando_imb}                                                           & $L_{\mathrm{DWB}}=-\frac{1}{n} \sum_{i=1}^{n} \sum_{j=1}^{c} w_{j}^{\left(1-p_{i j}\right)} y_{i j} \log \left(p_{i j}\right)-p_{i j}\left(1-p_{i j}\right)$               & CV                                                    & \begin{tabular}[c]{@{}c@{}}Cyber intrusion detection\\ Skin lesion diagnosis \\ (Imbalanced)\end{tabular}                                         & \begin{tabular}[c]{@{}c@{}}pred and \\ Category \\ proportion \end{tabular}                                                                  & Hard-first                                                                                    & Sample      \\ \hline
\end{tabular}}
\label{tab:plain}
\end{table*}
Table~\ref{tab1} lists some of the typical weighting schemes in previous literature. The core of a weighting scheme is its weighting function for the input samples. The weighting functions can be sample-wise, category-wise, or their mixtures. According to the priority mode, existing weighting functions can be easy-first and hard-first.  The application scenarios (i.e., standard, imbalanced, and noisy) of these functions are also presented. The hyper-parameters in most functions are nearly fixed during training, whereas they are dynamic in SPL~\cite{MPawanKumar04}.

Each weighting scheme in Table~\ref{tab1} can only implement one mode. Their corresponding modes are selected based on a (partial) particular view of the data characteristics. Focal loss~\cite{TsungYiLin03} is inspired by the observation that ``easy samples occupy more than hard ones in object detection data sets". SuperLoss~\cite{ThibaultCastells21} downweights the contribution of samples with a large loss and is effective when the training data is corrupted by noise.  LOW~\cite{CarlosSantiagoa19} forces the model to focus on less represented or more challenging samples and works well for imbalanced data. To overcome the positive-negative imbalance, ASL~\cite{EmanuelBenBaruch10} reduces the contribution of easy negative samples to the loss function. DWB~\cite{Fernando_imb} assigns higher
weights to hard to train instances based on the prediction probability of ground-truth class and the category proportion. There are also some studies inspired by other clues such as the human learning mechanism. For example, Curriculum learning~\cite{YoshuaBengio15} is motivated by human learning that easy samples should be learned first.
\subsection{The measurement criteria for learning difficulty}
Learning difficulty is an important attribute of samples. To answer the ``easy-or-hard" question, the measurements for learning difficulty should be clarified at first. Learning difficulty depends on various factors which are related to the samples' characteristics, including data quality~\cite{WonyoungShin79, BuyuLi11, Verleysen91}, sample neighbors~\cite{JingfengZhang16, DeliChen35}, and category distribution~\cite{CarlosSantiagoa19, YinCui17, SalmanHKhan44}. For example, the lower the quality of a sample is, the larger the learning difficulty of the sample will be; the more heterogeneous samples in the neighborhood of a sample are, the larger the learning difficulty of the sample will be; the smaller a category is, the larger the learning difficulty of samples in this category.

Current criteria for learning difficulty measurement of methods in Table~\ref{tab1} include loss, gradient, category proportion, and margin. None of the above measurements can capture all the aforementioned factors. Table~\ref{tab2} lists the criteria and their corresponding typical methods. Loss is the most widely used criterion. Two recent methods including JTT~\cite{EvanZheranLiu20} and Truncated Loss~\cite{WenjieWang18} still use this criterion. In addition, loss-based measurements are more efficient than others, and they are effective in various learning tasks.

\begin{table*}[]
\caption{Statistics of measurement criteria}
\label{tab2}
\centering
\begin{threeparttable}
\resizebox{1.8\columnwidth}{!}{
\begin{tabular}{|c|c|c|c|}
\hline
{Criterion}   & {Method}                                                                                                                                                                      & {Number} & {Scenario}\\ \hline
Loss (pred)\tnote{1}& \makecell[c]{SPL\_Binary (2010), SPL\_Log (2014), Cost-sensitive SPL (2016),\\ Focal Loss (2017), QFL (2020), ASL (2020), SuperLoss (2020),\\ FOCI (2020), Truncated Loss (2021), JTT (2021) } & {10}  & \makecell[c]{Standard, Noise, Imbalance}     \\ \hline
Gradient                           & LOW (2021)                                                                                                                                                                             & 1  & Imbalance                             \\ \hline
Category proportion                & Class-balance (2019), G-RW (2021)                                                                                                                                                                 & 2 & Imbalance                               \\ \hline
Margin                             & GAIRAT (2021)                                                                                                                                                                                      & 1  & Standard                            \\ \hline
\end{tabular}}

\label{tab:plain}
\begin{tablenotes}
     \item[1] \footnotesize{The loss and pred are considered to be the same criterion because of $l=-log(p)$.}
   \end{tablenotes}
\end{threeparttable}
\end{table*}

\section{Preliminary answers}

Existing studies explicitly or implicitly give partial answers to the ``easy-or-hard" question on a specific view or scenario. Few studies have attempted to thoroughly discuss the applicable/inapplicable scenarios for different priority modes. 

Based on partial observations/conclusions of existing studies, the following clues for the investigated question are summarized.
\begin{itemize}
\item The weights for noisy samples should be reduced making the model less disturbed by noise~\cite{ThibaultCastells21, WenjieWang18}. 
In other words, the easy-first mode will be more effective on training sets with heavy noise.
\item If easy samples are excessive, the hard-first mode is preferred. For example, in the application of Focal loss~\cite{TsungYiLin03} in object detection, hard samples are assigned (relatively) high weights. In imbalance learning, head categories are relatively easy and thus easy samples are excessive when a high imbalance exists~\cite{CarlosSantiagoa19}. 
\item According to human learning mechanisms, easy samples should be learned first. 
\end{itemize}

The above preliminary answers are far from satisfactory for the ``easy-or-hard" question because there are still some deep concerns:
\begin{itemize}
\item[(i)] Are there any other possible priority modes in addition to the easy-first and hard-first modes?
\item[(ii)] The second answer listed above refers to the difficulty distribution of training samples. What is the relationship between the difficulty distribution and the priority mode?
\item[(iii)] In nearly all existing studies, the priority mode is fixed during the training procedure. Can the priority mode be changed during the entire training process?
\item[(iv)] When there is no information for factors such as noises and difficulty distributions discussed in the preliminary answers, is there an effective solution to assign weights on training samples? 
\end{itemize}

To solve the four subproblems listed above, both theoretical and empirical investigations are conducted simultaneously. The theoretical and experimental conclusions give more comprehensive answers to the four subproblems and thus the investigated question.

\section{Theoretical investigation}
A general theoretical framework is lack for the pursuing of the learning difficulty-based weights in existing studies. To conduct theoretical analysis, a general optimized objective function is firstly proposed. 
\subsection{A general optimized objective function}
Learning difficulty-based weighted loss is defined in Eq.~(1):
\begin{equation}
{\mathcal{L}_{d}} = \frac{1}{N}\sum\nolimits_i {{w_i}{l_i}},
\end{equation}
where $w_i$ depends on $d_i$. The weights adjust the difficulty distribution of the training samples. Let $d(x)$ be the learning difficulty of $x$. Let $P_{opt}[d(x)]$ and $P_{tr}[d(x)]$ be the optimal and real difficulty distributions of the training data, respectively. Let $p_{opt}[d(x)]$ and $p_{tr}[d(x)]$ be the densities of $d(x)$ for the optimal and real difficulty distributions of the training data, respectively. Let $\widetilde{P}_{tr}[{d(x)}]$ be the new distribution on samples with weights and $\widetilde{p}_{tr}[d(x)] \propto w[d(x)]{p}_{tr}[d(x)]$, where $\widetilde{p}_{tr}[d(x)]$ is the density of $\widetilde{P}_{tr}[{d(x)}]$. Theoretically, the optimal weights can be pursued with the following objective function:
\begin{equation}
\min\limits_{{w}}
KL(\widetilde{P}_{tr}[{d(x)}]||P_{opt}[{d(x)}]),
\end{equation}
where $KL$ is the Kullback–Leibler divergence~\cite{TimvanErven88} between two distributions. According to Eq.~(2), the optimal weights for a training sample $x$ is
\begin{equation}
w^*{[d(x)]} = \frac{p_{opt}[{d(x)}]}{p_{tr}[{d(x)}]}.
\end{equation}
In the following discussion, $x$ is omitted for brevity. Although $p_{tr}(d)$ can be approximately inferred, it is nearly impossible to infer ${p_{opt}(d)}$ for a given learning task. Nevertheless, if useful information (i.e., partial data characteristics cues) is available and reasonable assumptions are given, an appropriate value for $w^*{(d)}$ can be obtained. For example, if ${p_{opt}(d)}$ is assumed to be uniform and the ``category 
proportion" is used as the difficulty measure, the reciprocal of category proportion is the optimal weight for each training sample.

\subsection{Investigation for Subproblems (i) and (ii)}
\begin{figure*}[t]
    \centering
    \setlength{\abovecaptionskip}{-0.15cm}  
    \includegraphics[width=0.85\linewidth]{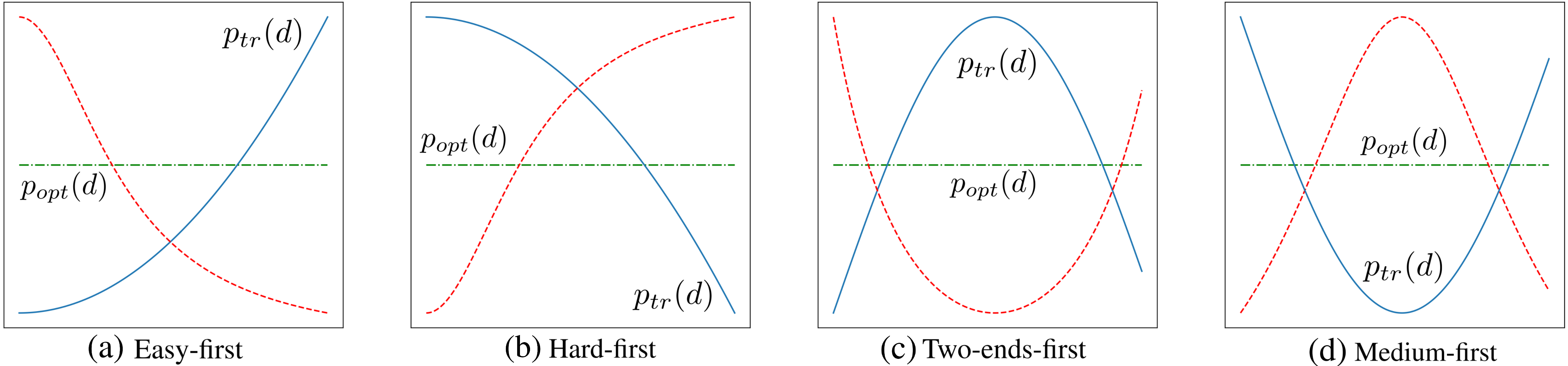} 
    \caption{Different difficulty distributions of the training data and the whole sample space. (The green line represents the difficulty distribution of the entire space which is assumed to be uniform, and the blue line represents the difficulty distribution of the training data. The red line is the optimal weight curve.)}\label{fig1}
\end{figure*}
Let $\tau(d)$ be $p_{opt}(d)/p_{tr}(d)$. According to Eq.~(3), four typical cases can be obtained and discussed as follows:

(1) $\tau(d)$ decreases on $d$, which means that hard samples are excessive in training. In this case, samples with low difficulties should be assigned with large weights. Alternatively, ``easy-first" is adopted. Fig.~\ref{fig1}(a) illustrates this case. $p_{opt}(d)$ is assumed to be uniform and the conclusion is also valid under any other distribution.

(2) $\tau(d)$ increases on $d$, which means that the easy samples are excessive in training. In this case, samples with high difficulties should be assigned with large weights. Alternatively, ``hard-first" is adopted. Fig.~\ref{fig1}(b) illustrates this case.

(3) $\tau(d)$ decreases at first and then increases on $d$, which means that medium-difficult samples are excessive in training. In this case, samples with low and high difficulties should be assigned with large weights. Alternatively, ``two-ends-first" is adopted. Fig.~\ref{fig1}(c) illustrates this case.

(4) $\tau(d)$ increases at first and then decreases on $d$, which means that both easy and hard samples excessive in training. In this case, samples with medium difficulties should be assigned with large weights. Alternatively, ``medium-first" is adopted. Fig.~\ref{fig1}(d) illustrates this case.

Based on the above analysis, the following answers are obtained:

\begin{itemize}
\item For Subproblem (i), at least two typical priority modes exist, namely, medium-first and two-ends-first. In deed, these two modes have been utilized in existing studies. For example, GHM~\cite{BuyuLi11} decreases the weight of over-hard (noise) samples on the basis of hard-first mode as Focal Loss is sensitive to noise. Thus, in our study, GHM is under the medium-first mode. Yang et al.~\cite{JufengYang23} proposed the self-paced balance learning whose priority mode is a combination of
the easy-first and hard-category-first modes. This priority mode is an approximation of the two-ends-first. 
\item For Subproblem (ii), Eq.~(3) well reveals the relationship between the priority modes and the two learning difficulty distributions, namely, $P_{opt}(d)$ and $P_{tr}(d)$. 
\end{itemize}

\subsection{More analysis for Subproblem (ii) based on the bias-variance trade-off theory}
The bias-variance trade-off theory is an effective method to describe the generalization error of the model. In this section, this is used to show that the optimal model complexity can be changed by weighting the training data. Furthermore, it supports the relationship between the learning difficulty distribution and the priority mode. 

Let $T$ be a random training set and ${f}(x|T)$ be the trained model on $T$. The bias-variance trade-off is based on the following learning error~\cite{ZitongYang82}:
\begin{equation}\small
\begin{aligned}
Err &= {E_{x,y}}{E_T}\left[ {\left\| {y - f\left( {x\left| T \right.} \right)} \right\|_2^2} \right] \\ &= Bias^2 + Variance + {\delta _e} \\
&\approx BiasT + VarT.
\end{aligned}
\end{equation}
The bias-variance trade-off theory indicates that the bias and variance terms will respectively decrease and increase if the model complexity $c$ increases~\cite{PedroDomingos25}. Minimum learning error is achieved when the sum of the partial derivatives of two terms with respect to the model complexity $c$ is equal to zero~\cite{ScottFortmannRoe81}. In this study, training samples are divided into easy, medium, and hard according to their learning difficulties. Therefore, we divide the sample space into three corresponding regions, namely, $R_{easy}$, $R_{medium}$, and $R_{hard}$. Similar to Eq.~(7), the learning error in $R_{easy}$ is defined:
\begin{equation}
\begin{aligned}
Er{r_{easy}} &= {E_{(x,y) \in {R_{easy}}}}{E_T}\left[ {\left\| {y - f\left( {x\left| T \right.} \right)} \right\|_2^2} \right] \\
&\approx BiasT_{easy} + VarT_{easy}.
\end{aligned}
\end{equation}
Likewise, we can define the bias/variance terms for the $R_{medium}$ and $R_{hard}$ regions. Based on the bias-variance trade-off theory on the entire sample space, we propose the following assumption:

\textit{\textbf{Assumption 1}}: For all the three bias (e.g., $BiasT_{easy}$) and variance terms (e.g., $VarT_{easy}$) of $R_{easy}$, $R_{medium}$, and $R_{hard}$, the bias and variance terms are decreasing and increasing functions of the model complexity $c$, respectively. Both the partial derivatives of the bias and variance terms with respect to $c$ are increasing functions, respectively. 

According to Assumption 1, minimum learning error for each region is achieved when the sum of the partial derivatives of its bias and variance term with respect to $c$ equals to zero.

Let $c^{*}$ be the optimal model complexity for the entire sample space when the minimum of $Err$ in Eq.~(12) is attained. Likewise, let $c^{*}_{easy}$ and $c^{*}_{hard}$ be the optimal model complexities for $R_{easy}$ and $R_{hard}$, respectively. The following assumption is proposed:

\textit{\textbf{Assumption 2}}: $c^{*}_{easy} < c^{*} < c^{*}_{hard}$.

With Assumption 2, we have the following propositions. 

\textit{\textbf{Proposition 1}}: If weights higher than one are exerted on the samples in $R_{hard}$, and the weights in the other regions remain one, then the new optimal model complexity $c^{*}_{new}$ over the entire space will be larger than $c^{*}$. Alternatively, the complexity of the optimal model is increased.

\textit{\textbf{Proposition 2}:} If weights higher than one are exerted on samples in the $R_{easy}$, and the weights in other regions remain one, the new optimal model complexity $c^{*}_{new}$ over the entire space will be smaller than $c^{*}$. Alternatively, the complexity of the optimal model is decreased. 

A theoretical analysis for {Proposition 1} is shown in the Appendix. {Proposition 1} supports that when easy samples are excessive in a training set, the trained model will become quite simple and under-fitting. Thus, hard samples should be assigned high weights to increase the complexity of the final model. Alternatively, the hard-first mode should be selected. {Proposition 2} supports that when hard samples are excessive in a training set, the easy-first mode should be selected. These two propositions further support the relationship between the learning difficulty distribution and the priority mode that we have discussed above. 

\subsection{Investigation for Subproblem (iii)}

Subproblem (iii) concerns that whether the fixed way is always the optimal, 
as the priority mode is fixed in all existing studies. According to Eq.~(3), the priority mode depends on the function $\tau(d)$. If $\tau(d)$ is fixed during the training procedure, the priority mode should not be changed. For example, when ``category proportion" is used as the measurement~\cite{YinCui17}, $\tau(d)$ will be fixed during the training process.

$\tau(d)$ depends on both ${p_{opt}(d)}$ and ${p_{tr}(d)}$. Although ${p_{opt}(d)}$ is unknown, it is fixed in training process. Therefore, the answer to Subproblem (iii) is subject to ${p_{tr}(d)}$. As described in Section~II(B), the learning difficulty is measured by heuristic factors such as loss and gradient in existing studies. These factors are varied during the training procedure, so ${p_{tr}(d)}$ is not fixed in the training process. If the variation of ${p_{tr}(d)}$ is large, then $\tau(d)$ will change greatly. Theoretically, the priority mode will be changed if ${p_{tr}(d)}$ changes drastically. For example, when loss is used as the measurement of learning difficulty, in the early stage of training, there will be numerous samples with large loss as shown in Fig.~1(a). With the model trained better, easy samples will dominate the training set if the training set has no great deviations (e.d., heavy noise) as shown in Fig.~1(b). Thus, during the training process, 
``easy-first" should be adopted in the initial training stage and ``hard-first" should be adopted gradually.

Based on the above analysis, an answer to Subproblem (iii) can be obtained that the priority mode is not absolutely fixed during training and it should be changed if ${p_{tr}(d)}$ changes greatly.

\subsection{Investigation for Subproblem (iv)}
If there is no prior knowledge, theoretical clues, or empirical observations, the self-paced mechanism may be helpful as they are inspired by the human learning process. However, this solution is not effective in all scenarios especially for imbalanced data~\cite{XiaoxiaWu31, GuyHacohen88}. To explore a more general and effective solution (answer) to Subproblem (iv), a data-adaptive strategy is proposed in this study. Specifically, a weighting function is defined with the condition that all the four typical modes described in Section IV-B can be realized by this function when different hyper-parameters are chosen.

Let $\Lambda$ be the hyper-parameter(s) of a weighting function. According to the four typical weight curves shown in Fig.~1, a weighting function $w(d; \Lambda)$, which can achieve the four priority modes, should satisfy the following three conditions:

(1) ${w}(d; \Lambda) \ge 0$;

(2) A hyper-parameter is required which can make a horizontal shift for $d$, which ensures that any segment of the entire curve can be taken. As a consequence, different priority modes can be implemented.

(3) $w(d; \Lambda)$ should contain  a local maximum and a local minimum under some certain values of the hyper-parameters;

Let $\alpha$ and $\gamma$ be the two hyper-parameters. There are various functions satisfying the above requirements, including trigonometric function (e.g., $sin(d+\alpha)$ and $cos(d+\alpha)$) and polynomial function (e.g., $(\alpha-d)^{\gamma}(\alpha+d)^{\gamma}$). 
Inspired by Focal loss, we propose a flexible weight function (FlexW) defined as follows:
\begin{equation}
\begin{aligned}
{w_i} = (d_{i} + \alpha)^{\gamma}e^{-\gamma(d_{i} + \alpha)}
\end{aligned}.
\end{equation}

In this study, $d$ is also approximated by $1-p$~\cite{TsungYiLin03}. If $\alpha \ge 0$, Condition (1) is satisfied. Condition (2) is also obviously satisfied. We analyze that whether Condition (3) is satisfied. To simplify the form of FlexW, let $t = 1 - p + \alpha$. Thus, the weighting function becomes
\begin{equation}
\begin{aligned}
f(t) = t^{\gamma}e^{- \gamma t}
\end{aligned}.
\end{equation}
Taking the derivative of Eq.~(7), we obtain
\begin{equation}
\begin{aligned}
f^{'}(t) = \gamma t^{\gamma - 1}e^{- \gamma t}(1-t)
\end{aligned}.
\end{equation}
\begin{figure}[t] 
    \centering
    \includegraphics[width=0.65\linewidth]{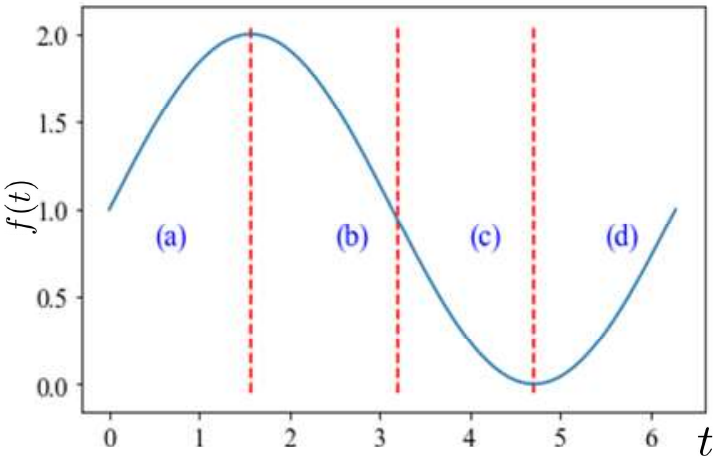} 
    \caption{The curve of $f(t)$ when $\gamma = 2$.}
    \label{fig3}
\end{figure}
Regardless of the value of $\gamma$, the function either has both a local maximum and a local minimum or only contains a local maximum or a local minimum. Thus, Condition (3) is satisfied. 

Fig.~2 shows the curve of the FlexW function when $\alpha=0$ and $\gamma = 2$. If the segment (a) is selected, then the ``hard-first" priority mode is implemented; if the  segment (b) or (c) is selected, then ``easy-first" priority mode is implemented; if the segments (a) and (b) are both selected, then the ``medium-first" priority mode is implemented; if the segments (c) and (d) are both selected, then the ``two-ends-first" priority mode is implemented. Considering that $t \in [0,1]$, which segment(s) is/are selected depends on the horizontal shift hyper-parameter $\alpha$. When $\gamma$ is set to zero, the weights of all samples are equal.


When different values of $\alpha$ and $\gamma$ are chosen, different priority modes can be produced by FlexW. Fig.~3 shows weight curve examples including ``easy-first" (Fig.~3(a)), ``medium-first" (Fig.~3(b)), ``hard-first" (Fig.~3(c)) and ``two-ends-first" (Fig.~3(d)) when different values of $\alpha$ and $\gamma$ are set. Therefore, we only need to change the values of $\alpha$ and $\gamma$ of FlexW instead of the entire weighting function when facing different learning tasks. Our experimental evaluation of different scenarios shows the considerable flexibility of our proposed FlexW.

\begin{figure*}[t]
    \centering
    \includegraphics[width=1\linewidth]{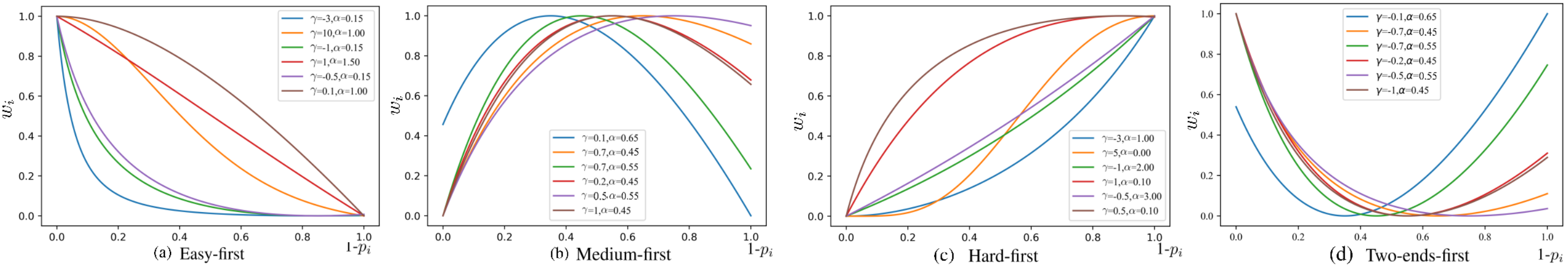} 
    \caption{The weight curves of FlexW for the four priority modes when different hyper-parameters are set.}\label{fig2}
\end{figure*}
Based on the above analysis, an answer to Subproblem (iv) is obtained: when there is no prior knowledge or theoretical clues for the learning task, a weighting function (e.g., FlexW) which can achieve four priority modes can be used. The optimal mode (i.e., the optimal hyper-parameters $\Lambda$) can be selected based on the tuning of the hyper-parameters on validation data. 

\begin{table*}[t]
\caption{Test accuracies (\%) of the competing methods under flip noise. The best and the second best results are bold and underlined. * means that the SPL manner is combine into FlexW.}
\label{tab3}
\centering
\resizebox{2\columnwidth}{!}{
\begin{tabular}{lccccccccccc|cc|c}
\Xhline{0.12em}
Data set                  & Noise & Baseline   & Reed Hard  & SPL\_Binary & SPL\_Log   & Focal loss & S-model    & \begin{tabular}[c]{@{}l@{}}Co-teaching\end{tabular} & D2L        & \begin{tabular}[c]{@{}l@{}}Fine-tuning\end{tabular} & MentorNet  & \begin{tabular}[c]{@{}l@{}}FlexW (hard-first)\end{tabular} & \begin{tabular}[c]{@{}l@{}}FlexW (easy-first)\end{tabular}& 
\begin{tabular}[c]{@{}l@{}}FlexW (easy-first*)\end{tabular} \\ \Xhline{0.05em}
\multirow{2}{*}{CIFAR10}  & 20\%  & 76.83±2.30 & 88.28±0.36 & 87.03±0.34  & \underline{89.50±0.48} & 86.45±0.19 & 79.25±0.30 & 82.83±0.85                                             & 87.66±0.40 & 82.47±3.64                                             & 86.36±0.31 & 82.25±0.36                                                   & 89.28±0.15 
& \textbf{90.96±0.12}                                          \\
                          & 40\%  & 70.77±2.31 & 81.06±0.76 & 81.63±0.52  & 84.01±0.51 & 80.45±0.97 & 75.73±0.32 & 75.41±0.21                                             & 83.89±0.46 & 74.07±1.56                                             & 81.76±0.28 & 75.45±0.64                                                   & \underline{84.56±0.35}
                          & \textbf{85.64±0.11}                                          \\ \Xhline{0.05em}
\multirow{2}{*}{CIFAR100} & 20\%  & 50.86±0.27 & 60.27±0.76 & 63.63±0.30  & \underline{63.82±0.27} & 61.87±0.30 & 45.45±0.25 & 54.13±0.55                                             & 63.48±0.53 & 56.98±0.50                                             & 61.97±0.47 & 53.55±0.25                                                   & 63.76±0.12 
& \textbf{65.48±0.82}                                          \\
                          & 40\%  & 43.01±1.16 & 50.40±1.01 & 53.51±0.53  & 53.20±0.11 & 54.13±0.40 & 43.81±0.15 & 44.85±0.81                                             & 51.83±0.33 & 46.37±0.25                                             & 52.66±0.56 & 44.88±0.58                                                   & \textbf{55.56±0.24} 
                          & \underline{55.50±0.25}                                          \\ 
\Xhline{0.12em}
\end{tabular}}
\label{tab:plain}
\end{table*}
\begin{table*}[htbp]
\caption{Test accuracies (\%) of the competing methods under uniform noise.}
\label{tab4}
\centering
\resizebox{2\columnwidth}{!}{
\begin{tabular}{lccccccccccc|cc|c}
\Xhline{0.12em}
Data set                  & Noise ratio & Baseline   & Reed Hard  & SPL\_Binary & SPL\_Log & Focal loss & S-Model    & Co-teaching    & D2L        & Fine-tuning & MentorNet  & FlexW (hard-first)& FlexW (easy-first)& FlexW (easy-first*)\\ \Xhline{0.05em} 
\multirow{2}{*}{CIFAR10}  
                          & 40\%  & 68.07±1.23 & 81.26±0.51 &
                          86.41±0.29  & 77.50±0.50 &
                          75.96±1.31 &
                          79.58±0.33 &
                          
                          74.81±0.34  & 
                          85.60±0.13 &  80.47±0.25  & 87.33±0.22 & 76.28±0.35                    & \underline{87.64±0.24}                                       & \textbf{88.15±0.22}                                                         \\
                          & 60\%  & 53.12±3.03 & 73.53±1.54 &53.10±1.78  & 53.40±0.38 &51.87±1.19 & -          & 73.06±0.25  &  68.02±0.41 &  78.75±2.40  & \textbf{82.80±1.35} & 69.63±1.08                        & 79.82±0.54                                           & \underline{81.87±1.23}                                                         \\ \Xhline{0.05em} 
\multirow{2}{*}{CIFAR100} 
                          & 40\%  & 51.11±0.42 & 51.27±1.18 &
                          55.11±0.75  & 54.94±0.21 &
                          51.19±0.46 &42.12±0.99 & 46.20±0.15  &  52.10±0.97 &  52.49±0.74  & \textbf{61.39±3.99} & 51.85±0.55                        & \underline{58.48±0.34}                                         & 57.72±0.36                                                        \\
                          & 60\%  & 30.92±0.33 & 26.95±0.98 &36.56±0.57  & 37.17±0.32 & 27.70±3.77 & -          & 35.67±1.25  &  41.11±0.30 & 38.16±0.38  & 36.87±1.47 & 35.22±1.55                     & \underline{41.40±0.75}                                      & \textbf{42.50±0.87}                                                         \\ \Xhline{0.12em} 
\end{tabular}}
\label{tab:plain}
\end{table*}
\section{Experimental investigation}
Section VI performs theoretical investigation to the four subproblems listed in Section III. This section  conducts extensive experiments for various tasks under different scenarios. Even tough Our theoretical analyses are adapted to all difficulty measurements, loss is used as the measurement of learning difficulty as with most methods shown in Table~II in our experiments. 

First, we consider two typical factors which influence the samples' learning difficulties including noise and imbalance. Second, different difficulty distributions of training data set are investigated in object detection. Third, the performance of ``varied modes" is studied. Fourth, we verified that when no prior knowledge or theoretical clues exist, weighting schemes (e.g., FlexW) that can achieve four priority modes should be adopted. 
\begin{figure}[b] 
    \begin{center}
    \setlength{\belowcaptionskip}{-0.15cm}   
    \includegraphics[width=1\linewidth]{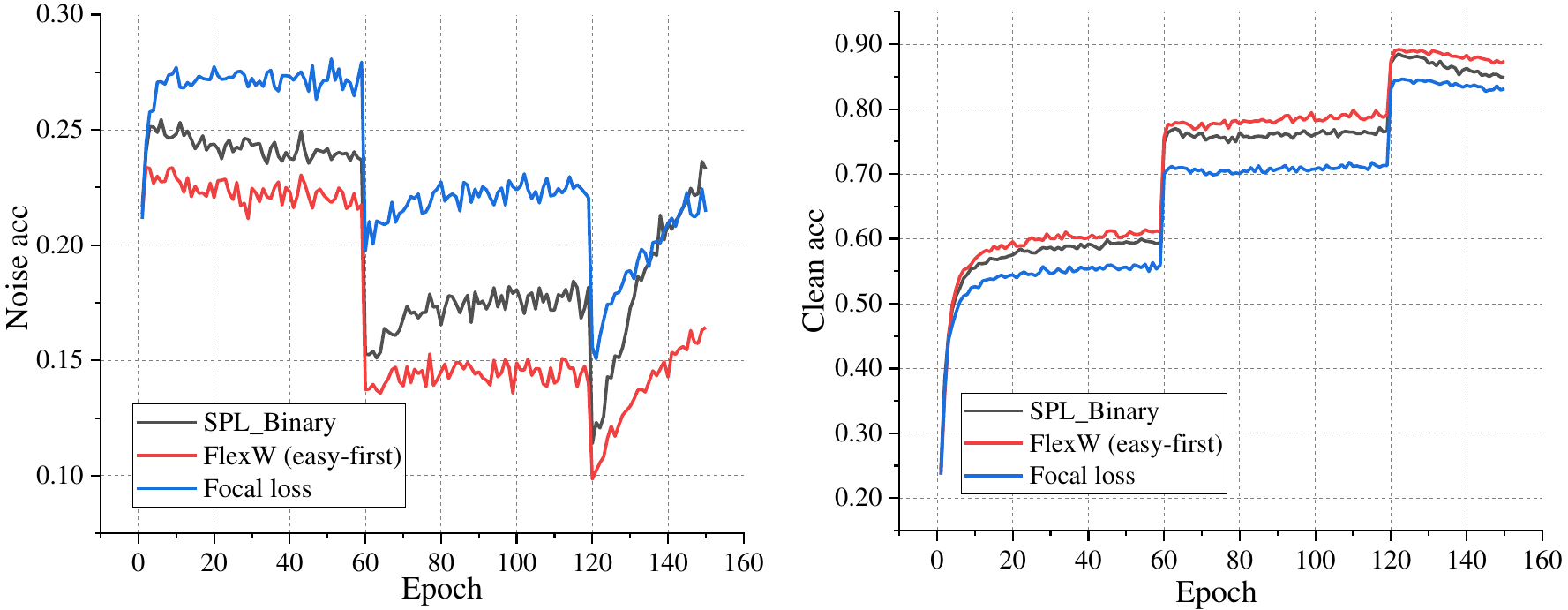} 
    \vspace{-0.22in} 
    \caption{Accuracies of the three methods on noisy (left) and clean (right) samples under 40\% flip noise.}
    \label{fig4}
    \end{center}
\end{figure}
\subsection{Image classification with noisy labels}
\subsubsection{Experimental settings}
Two benchmark data sets, namely, CIFAR10 and CIFAR100~\cite{AlexKrizhevsky27}, are used. Each sample is a $32 \times 32$ image from 1 out of 10 or 100 categories. Flip and uniform label noises are simulated following the manners in~\cite{JunShu28}. Wide ResNet-28-10 (WRN-28-10)~\cite{SergeyZagoruyko29} and ResNet-32~\cite{KaimingHe30} are adopted for the flip and uniform noises, respectively. Each experimental run is repeated five times with different seeds for parameter initialization and label noise generation. 

The comparison methods include Baseline which refers to the basic classifier network with CE loss and the advanced learning methods including Reed~\cite{ScottReed60}, S-Model~\cite{JacobGoldberger61}, SPL~\cite{MPawanKumar04,LuJiang06}, Focal loss~\cite{TsungYiLin03}, Co-teaching~\cite{BoHan58}, D2L~\cite{XingjunMa59}, MentorNet~\cite{LuJiang41}, and Fine-tuning~\cite{JunShu28} which refers to fine-tuning the result of Baseline on the meta-data with clean labels to further enhance its performance.

In this experiment, the networks are trained using SGD with a momentum $0.9$, a weight decay $5 \times e^{-4}$, and an initial learning rate $0.1$. At the 60th, 120th, and 160th epochs, the learning rate is reduced to one-fifth of that in the previous epoch. The values of the batch size and epoch are set to 32 and 200, respectively. The values of $\gamma$ and $\alpha$ for easy/hard/medium/two-ends-first modes of FlexW are searched in $\{-0.6, -0.5, -0.4, -0.2\} \times \{0.1, 0.2, 0.3\}$ / $\{0.2, 0.4, 0.5, 0.6\} \times \{0.1, 0.2, 0.3\}$ / $\{0.2, 0.4, 0.5, 0.6\} \times \{0.4,0.6,0.8\}$ / $\{-0.6, -0.5, -0.4, -0.2\} \times \{0.4, 0.6, 0.8\}$, respectively. During the experiments, the weights are normalized.
\subsubsection{Results}
Adding noise will change the difficulty distribution, resulting in excessive hard samples in the training set relative to the entire space. To analyze the performances of the hard-first and easy-first modes on noise data, the specific accuracies of SPL\_Binary (easy-first), Focal loss (hard-first), and FlexW (easy-first) on noisy and clean samples are analyzed which is shown in Fig.~\ref{fig4}. The schemes with the easy-first mode (including SPL\_{Binary} and FlexW (easy-first)) have lower accuracies on noisy samples compared with the hard-first method (i.e., Focal loss)  before 140 epochs as shown in the left figure. 
From the right figure, the easy-first methods including SPL\_Binary and FlexW (easy-first) consistently outperform Focal loss which is under the hard-first mode on clean samples. Therefore, the easy-first methods are less affected by noise and make the model pay more attention to clean samples.

Four priority modes implemented by FlexW are experimented on CIFAR10 under 20\% and 40\% flip noises. The results are shown in Table~\ref{tab5}. It shows that the medium-first mode also achieves good performances on noise data. Therefore, the easy/medium-first modes are more suitable than hard-first ones on noisy data. Under different noise rates, we compare various advanced methods, as shown in Tables~\ref{tab3} and~\ref{tab4}. The easy-first methods (i.e., SPL and FlexW~(easy-first)) perform better than the hard-first ones (i.e., Focal loss, FlexW~(hard-first)) on noisy data.
Adding SPL manner into FlexW can further improve the performances in some cases. The weighting function of FlexW with SPL manner is
\begin{equation}
\begin{aligned}
{w_i} = \left\{ \begin{array}{l}
{(1 - {p_i} + \alpha )^\gamma }{e^{ - \gamma (1 - {p_i} + \alpha )}},\quad {l_i} \le \lambda \\
0,\quad {l_i} > \lambda
\end{array} \right.
\end{aligned}.
\end{equation}

\begin{table}[t]
\caption{The performances of different parameter settings on CIFAR10 under 20\% and 40\% flip noises. }
\label{tab5}
\centering
\begin{tabular}{lllll}
\Xhline{0.1em}
Noise rate & Easy-first & Medium-first & Hard-first & Two-ends-first\\ \Xhline{0.05em}
20\% & 89.28+0.15 & 89.17+0.24   & 82.25+0.36 & 80.13+0.52\\
40\% & 84.56+0.35 & 85.81+0.45   & 75.45+0.64 & 75.06+0.38 \\ \Xhline{0.1em}
\end{tabular}
\end{table}
\begin{figure}[t]
    \centering
    \includegraphics[width=0.85\linewidth]{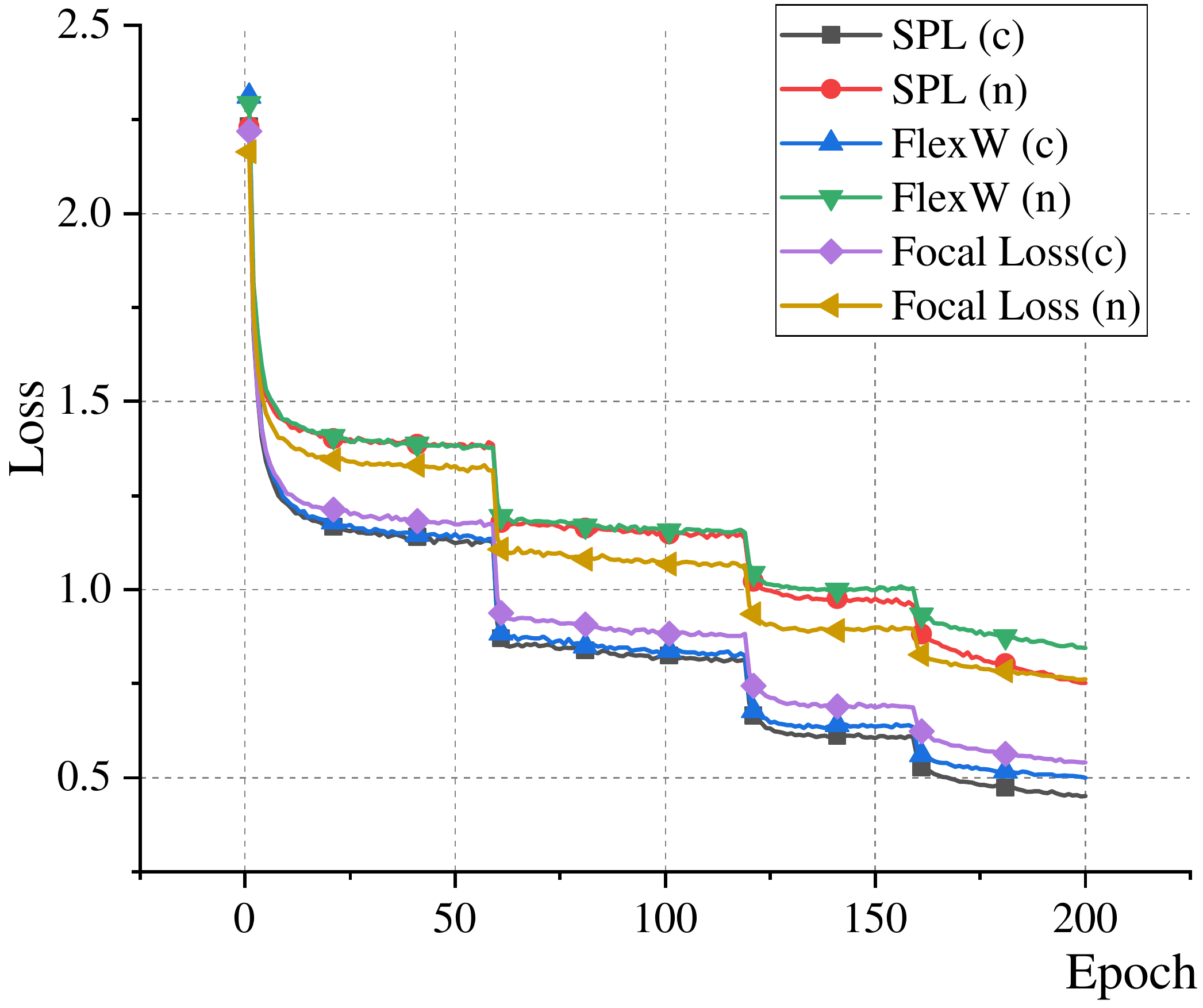} 
    \vspace{-0.1in}
    \caption{The average losses of clean and noisy samples of the three methods including SPL\_binary, Focal loss and FlexW (easy-first). ``c" and ``n" are the abbreviations of ``clean'' and ``noise''.}
    \label{fig5}
\end{figure}
$\lambda$ is a hyper-parameter defined in SPL~\cite{MPawanKumar04}. The average losses of clean samples and noisy samples under 40\% flip noise are shown in Fig.~\ref{fig5}. The average loss of the noisy samples is always higher than that of the clean ones during the training process. It indicates that noisy samples are more difficult than clean ones. What's more, using loss as the criterion to distinguish clean and noisy samples is reasonable.

The experimental results are consistent with our theoretical analysis for Subproblem (ii) shown in Sections IV-B and IV-C (Proposition 2). 


\subsection{Image classification with imbalanced data sets}
\begin{figure}[b]
    \centering
    \includegraphics[width=0.49\linewidth]{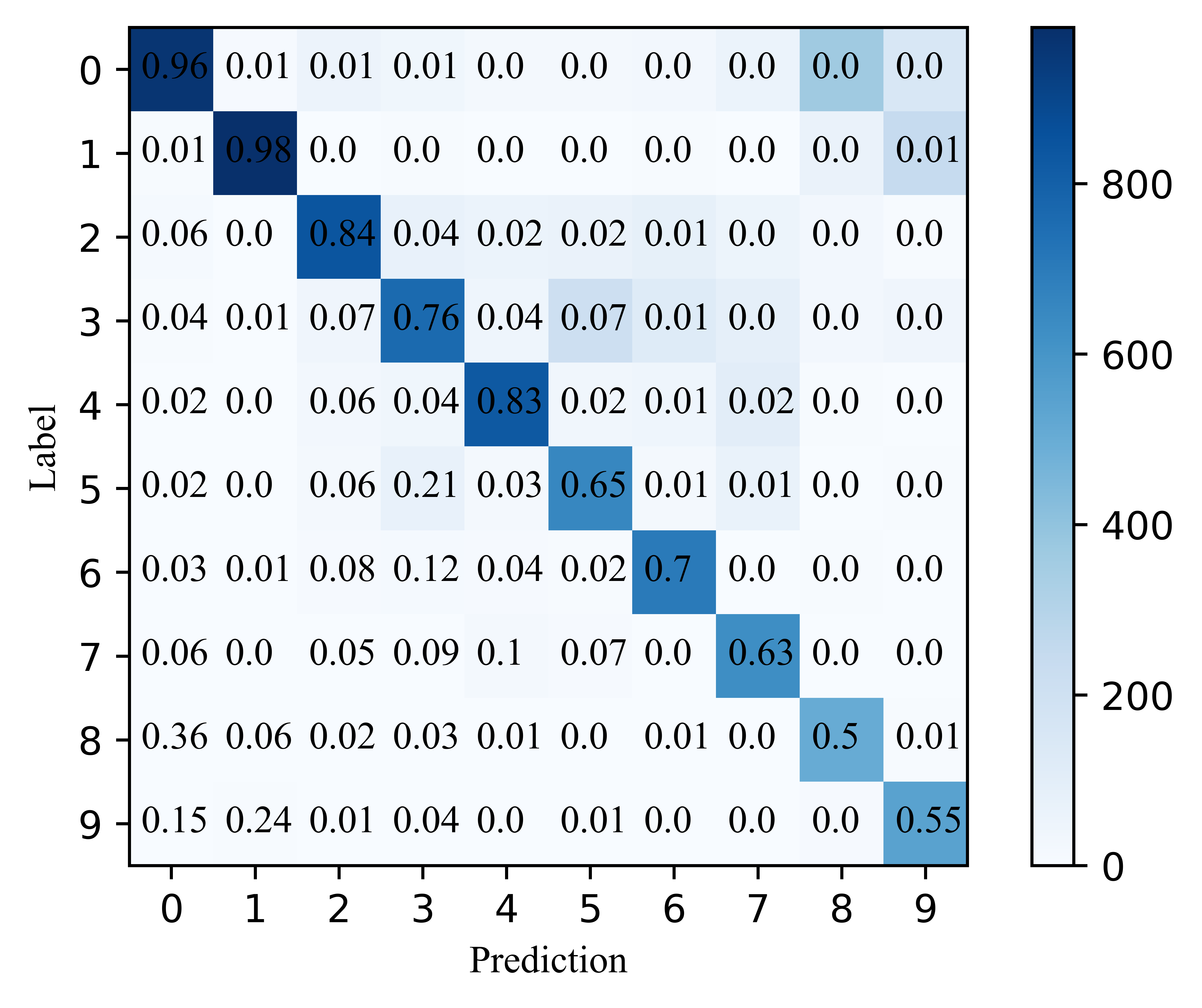} 
    \includegraphics[width=0.49\linewidth]{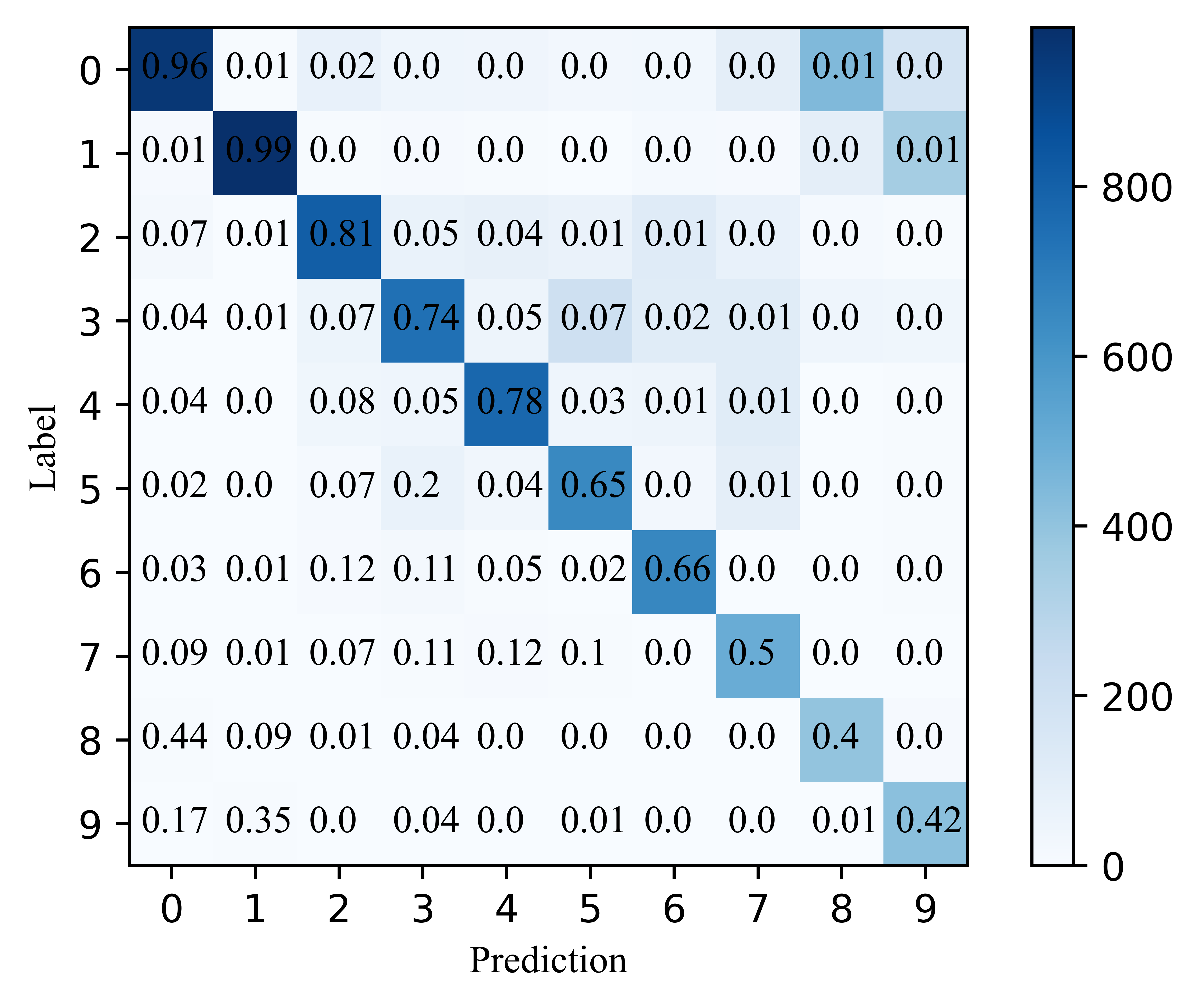} 
    \vspace{-0.22in} 
    \caption{Confusion matrices of the true labels and predictions obtained by FlexW  (hard-first) (left) and SPL\_Binary (right).}
    \label{confusion}
\end{figure}
\begin{figure*}[t]
    \centering
    \includegraphics[width=1\linewidth]{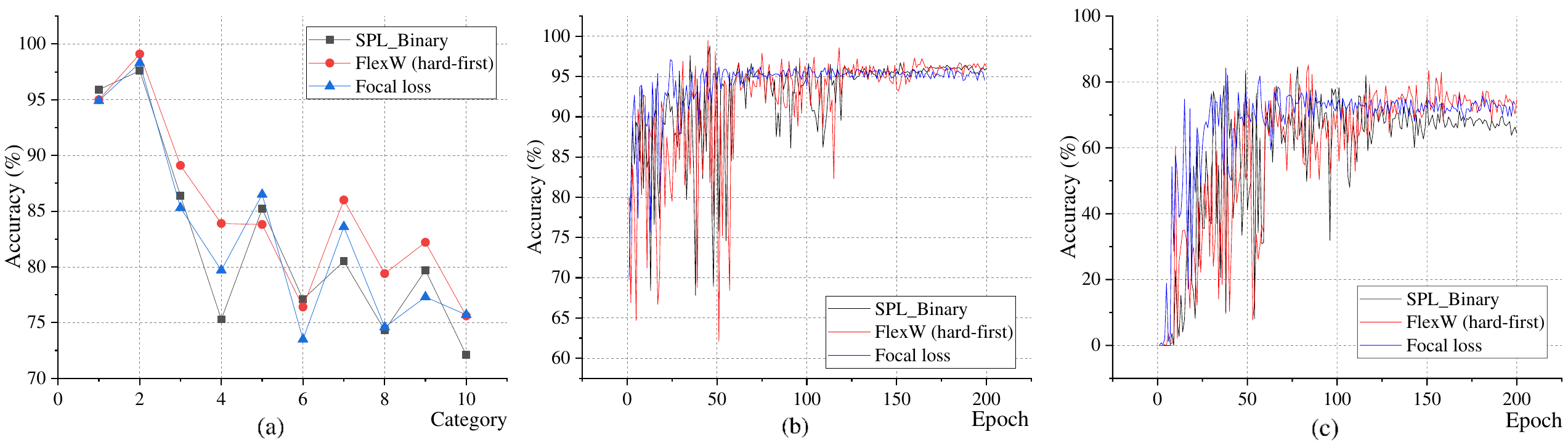} 
    \vspace{-0.3in} 
    \caption{(a) shows the accuracies of ten categories in their respective optimal epochs; (b) and (c) show the accuracies of Categories 1 and 10, respectively.}  
    \label{fig6}
\end{figure*}
\subsubsection{Experimental settings}
In this experiment, long-tailed versions of CIFAR benchmarks with different imbalance factors as defined by~\cite{YinCui17} are used. ResNet-32~\cite{KaimingHe30} is used as the basic model. The compared methods include the Baseline model which uses a cross-entropy loss, Focal loss~\cite{TsungYiLin03},  SPL~\cite{MPawanKumar04,LuJiang06}, Mix up~\cite{HongyiZhang63}, LDAM~\cite{KaidiCao64}, Class-balanced~\cite{YinCui17}, Equalised~\cite{JingruTan65}, L2RW~\cite{MengyeRen67} which leverages an additional meta-data to adaptively assign weights for training samples, and Class-balanced Fine-tuning~\cite{YinCui66} which means that the model is fine-tuned using the meta-data. Other experimental settings are the same as Section~V-A.
\subsubsection{Results}
For imbalanced training data, the proportion of easy samples is larger than that of the entire space shown in Fig~1(b). To study the performances of the hard-first and easy-first modes on imbalanced data, the accuracy for each category is analyzed when the imbalance factor equals to 20. Fig.~\ref{fig6}(a) indicates that methods under hard-first mode (i.e., FlexW(hard-first) and Focal loss) increase the accuracies of most tail categories compared with those under easy-first mode (i.e., SPL\_Binary). For example, FlexW increases the accuracies in tail categories 7, 8, 9, 10 and Focal loss increases the accuracies in tail categories 5, 7, 8, 10. Fig.~\ref{confusion} shows that compared with SPL\_Binary (easy-first), FlexW (hard-first) greatly improves the accuracies of tail categories when the imbalance factor equals to 100. Fig.~\ref{fig6}(c) shows that the methods of the hard-first mode significantly improve the accuracy of the last tail category. Since the first head category contains a large number of samples, all three methods perform well shown in Fig.~\ref{fig6}(b). Thus, for imbalanced data, the effective method should focus on improving the classification effect of the tail categories which can be achieved by the hard-first mode according to the above analysis.

\begin{table*}[b]
\caption{Test accuracies (\%) on imbalanced CIFAR10 and CIFAR100 with different imbalance factors. * means that the category-level scale parameter is combined into FlexW.}
\label{tab6}
\centering
\begin{tabular}{l|ccccc|ccccc}
\Xhline{0.1em}  
Data set                     & \multicolumn{5}{c|}{Long-tailed CIFAR10} & \multicolumn{5}{c}{Long-tailed CIFAR100} \\ \Xhline{0.05em}  
Imbalance factor                 & 200    & 100    & 50    & 20    & 10    & 200    & 100    & 50     & 20    & 10    \\ \Xhline{0.05em}  
CE (Baseline)                     & 65.68  & 70.36  & 74.81 & 82.23 & 86.39 & 34.84  & 38.32  & 43.85  & 51.14 & 55.71 \\
Focal loss\_$\gamma$=1             & 65.29  & 70.38  & 76.71 & 82.76 & 86.66 & 35.62  & 38.41  & 44.32  & 51.95 & 55.78 \\
Focal loss\_$\gamma$=0.5           & 64.00  & 70.33  & 76.72 & 82.89 & 86.81 & 35.00  & 38.69  & 44.12  & 51.10 & 55.70 \\
SPL\_Binary                      & 65.64  & 70.94  & 76.82 & 82.41 & 87.09 & 35.56  & 38.16  & 42.77  & 50.91 & 56.70 \\
SPL\_Log                         & 62.05  & 70.46  & 75.64 & 82.66 & 86.62 & 33.08  & 38.51  & 41.71  & 49.71 & 54.79 \\
Class-balance CE loss    & \underline{68.77}  & 72.68  & 78.13 & \underline{84.56} & 87.90 & 35.56  & 38.77  & 44.79  & 51.94 & 57.57 \\
Class-balance Fine-tuning & 66.24  & 71.34  & 77.44 & 83.22 & 83.17 & \textbf{38.66}  & 41.50  & 46.42  & 52.30 & 57.57 \\
Class-balance Focal loss   & 68.15  & 74.57  & \underline{79.22} & 83.78 & 87.48 & 36.23  & 39.60  & 45.21  & 52.59 & 57.99 \\
LDAM                       & 66.75  & 73.55  & 78.83 & 83.89 & 87.32 & 36.53  & 40.60  & 46.16  & 51.59 & 57.29 \\ 
Equalised                 & -      & 73.98  & -     & -     & -     & -      & \textbf{42.74}  & -      & -     & -     \\
Mixup                      & -      & 73.06  & 77.82 & -     & 87.10 & -      & 39.54  & 44.99  & -     & 58.02 \\
Meta-weight net            & 67.20  & 73.57  & 79.10 & 84.45 & 87.55 & 36.62  & 41.61  & 45.66  & \underline{53.04} & \underline{58.91} \\
L2RW                       & 66.25  & 72.23  & 76.45 & 81.35 & 82.12 & 33.00  & 38.90  & 43.17  & 50.75 & 52.12 \\
FlexW (easy-first)             & 64.82  & 70.48  & 76.36 & 82.45 & 86.56 & 34.84  & 38.51  & 43.42  & 50.45 & 56.47 \\ 
FlexW (hard-first)                & 68.74  & \textbf{75.41}  & 79.08 & 84.55 & \textbf{88.56} & 36.71  & 40.92  & \underline{46.32}  & 52.54 & 58.62 \\
FlexW (hard-first*)                & \textbf{69.40}  & \underline{75.33}  & \textbf{80.05} & \textbf{85.46} & \underline{88.50} & \underline{37.54}  & \underline{41.69}  & \textbf{47.18}  & \textbf{53.10} & \textbf{58.98} \\ \Xhline{0.1em} 
\end{tabular}
\label{tab:plain}
\end{table*}
Table~\ref{tab6} compares the performances of some advanced methods under different imbalance factors. Two typical hard-first methods (i.e., FlexW (hard-first) and Class-balance) perform well. The performances of hard-first methods (e.g., FlexW (hard-first) and class-balance) surpass those of the easy-first methods (e.g., SPL and FlexW (easy-first)). The scale parameter also improves the accuracies. The weighting function of FlexW combined with the scale parameter is
\begin{equation}
\begin{aligned}
{w_i} = c_{y_{i}}(1-p_{i} + \alpha)^{\gamma}e^{-\gamma(1-p_{i} + \alpha)}
\end{aligned},
\end{equation}
where $c_{y_{i}}$ is the category-level scale parameter. Furthermore, the performances of FlexW are ranking first or second in all cases. The performances of SPL are approaching those of Focal loss in some cases. It is because easy-first methods can improve the accuracies of the head categories. However, these methods further enlarge the gap between head and tail categories which is not expected.

\begin{figure*}[t]
    \centering
    \includegraphics[width=1\linewidth]{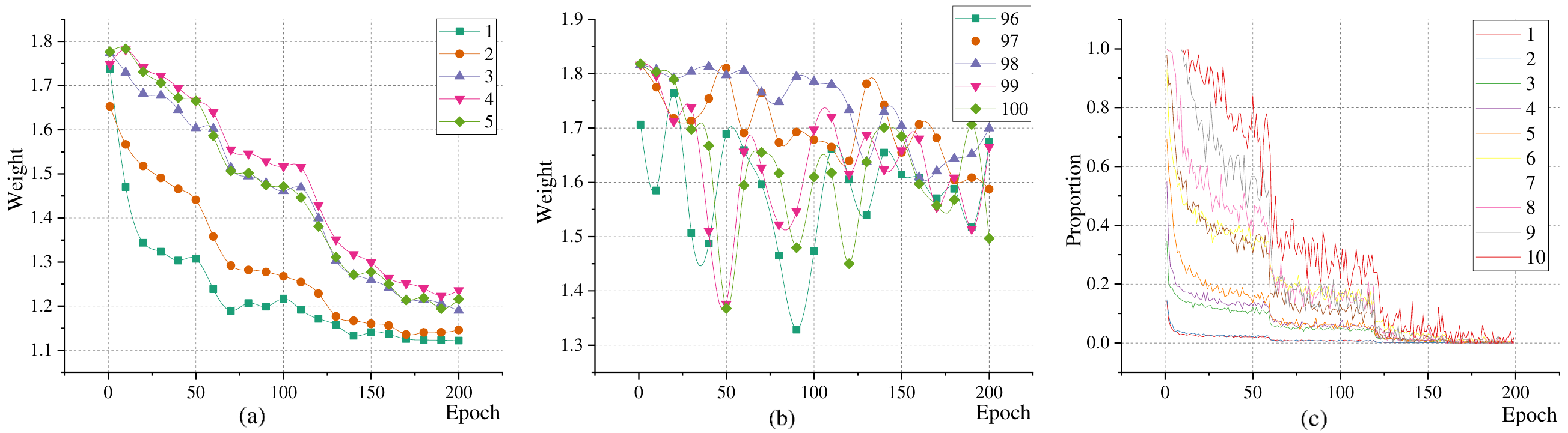} 
    \vspace{-0.3in} 
    \caption{(a) and (b) show the average weights of the first/last five head/tail categories on CIFAR100. (c) shows the proportion of hard samples contained in each category on CIFAR10. 
}
\label{fig7}
\end{figure*}

Figs.~\ref{fig7}(a) and (b) show the average weights of samples in the five head (a) and tail (b) categories, reflecting the contribution of samples in each category to the model. The weights of the five head categories drop quickly, whereas those of the tail categories remain high during the entire training process. It indicates hard-first mode increases the influence of the tail categories on the model which is exactly what we expect.

Fig.~\ref{fig7}(c) shows the proportion of hard samples (with $l_{i}$ $\ge \log{10}$) in each category. Tail categories have larger proportions of hard samples than head ones, which supports the common sense that samples in the tail categories are harder to learn well than those in the head on average.

The experimental results are in accordance with our theoretical analysis for Subproblem (ii) shown in Sections IV-B and IV-C (Proposition 1). When easy samples are excessive as shown in Fig.~\ref{fig1}(b), the trained model will be under-fitting. To appropriately increase the complexity of the trained model, the hard-first mode should be adopted.
\begin{table*}[b]
\caption{mAPs (\%) of the six weighting schemes on the four VOC data sets.}
\label{tab7}
\centering
\resizebox{1.8\columnwidth}{!}{
\begin{tabular}{lcccccc}
\Xhline{0.1em}  
Scheme & FL (hard-first) & FL (easy-first) & FlexW (hard-first) & FlexW (easy-first) & FlexW (medium-first) & FlexW (both-ends-first)\\ \Xhline{0.05em}  
VOC-e  & \underline{75.21}          & 66.96          & \textbf{76.84}              & 71.70      & 73.25  & 66.88      \\
VOC-h  & 66.62          & \underline{68.30}          & 67.67              & \textbf{69.25}  &67.64 &65.74       \\
VOC-m  & 55.74          & 62.36          & 60.14              & \underline{62.71}    &58.76 & \textbf{63.25}       \\ 
VOC-b  &   \underline{61.75}        &   57.72       &  61.43            &  59.54    &  \textbf{63.58}   &58.66   \\
\Xhline{0.1em} 
\end{tabular}}
\label{tab:plain}
\end{table*}
\subsection{Different difficulty distributions}
\subsubsection{Experimental settings}
Dense object detection is a typical application where the distribution of easy and hard samples is imbalanced. PASCAL VOC~\cite{EveringhamMark36,EveringhamMark37} is used in this experiment. The training set consists of VOC2007 and VOC2012 train+val with a total of 16,551 samples. As the training set contains excessive easy samples\cite{TsungYiLin03}, it is abbreviated as VOC-e. To investigate other data distribution cases, we compiled three training data sets based on VOC: data set with excessive hard samples (VOC-h), data set with excessive medium-difficult samples (VOC-m), and data set with both excessive easy and hard samples (VOC-b). All three artificially constructed training sets contain 8,000 images. For VOC-h, 7,000 images are from the images with the largest loss-conf values in the original VOC training set, and the remaining 1,000 images are randomly selected from training data except for the hardest 7,000 ones. VOC-m is composed of 8,000 images with moderate loss-conf values. VOC-b is composed of 4,000 images with the smallest loss-conf values and 4,000 images with the largest loss-conf values in the original VOC training set. VOC2007 test is used as the test set with a total of 4,952 samples. 

The YOLOv4~\cite{AlexeyBochkovskiy75} model, which utilizes Focal loss (FL) to calculate the loss of confidence, is used. The pre-trained weight trained by Darknet\cite{JosephRedmon89} is adopted. The optimizer we used is SGD where the momentum and weight decay are set to $0.9$ and $5 \times 10^{-4}$, respectively. The value of epoch is set to $50$ and the batch size is set to $4$. The initial learning rate is $1 \times 10^{-4}$, and the final learning rate is $1 \times 10^{-6}$. The value of the warm-up epoch is set to $2$. The settings for parameters in FlexW are the same as Section~V-A.
\subsubsection{Results}
In this experiment, we reveal an interesting fact that Focal loss can also implement the easy-first mode when its hyper-parameter $\gamma$ is negative. In Table~\ref{tab7}, we discuss the six weighting schemes (FL (easy-first), FL (hard-first), FlexW (easy-first), FlexW  (hard-first), FlexW (medium-first), FlexW  (both-ends-first)) for the four data sets.

The two hard-first schemes including FlexW (hard-first) and FL (hard-first) obtain better results on VOC-e which contains excessive easy samples. In contrast, when the data set has excessive hard samples, the easy-first methods (i.e., FlexW (easy-first) and FL (easy-first)) get better results. For VOC-m, FlexW (both-ends-first) gets the best performance and FlexW (easy-first) gets the second-best performance. For VOC-b, FlexW (medium-first) gets the best performance and FL (hard-first) gets the second-best result.

The experimental results are consistent with the four conclusions obtained in our theoretical analyses for Subproblems (i) and (ii) shown in Section~IV-B. Furthermore, FlexW achieves competitive performances on object detection tasks.

\subsection{The performances on varied priority modes}
As mentioned in Section~IV-D, if the difficulty distribution of the training set changes greatly during the training process, the corresponding priority mode should also be changed. When loss is used as the measurement of learning difficulty, the difficulty distribution of the training set will change during the training process. Using FlexW, the varied priority modes are investigated on imbalanced data. In the early training stages, most samples have high loss values, so that the easy-first mode should be adopted. With the model training better, easy samples will be excessive because hard samples are mostly samples in tail categories. Thus, the hard-first mode should be leveraged to improve the performance of the tail categories in later periods. 

Tables~\ref{tabix9} and \ref{tabix10} show the performances of FlexW with varied modes during training. ``Varied modes" used here means that the priority mode of easy-first is used in the first 100 epochs and hard-first is used in the rest of the epochs. The results show that the ``varied modes" achieves better performances compared with the fixed priority mode in some cases which is mainly because their difficulty distribution changes greatly during the training process as analyzed in Section~IV-D.

The experimental results are in accordance with the answer for Subproblem (iii). The priority mode does not need to remain fixed. The variation of the priority modes during the training process may achieve better results in some cases. 
\begin{table}[t]
\centering
\caption{Accuracies (\%) under the hard-first mode and varied modes of FlexW on CIFAR100. }
\label{tabix9}
\begin{tabular}{lccccc}
\Xhline{0.1em}
Imbalance factor      & 200            & 100            & 50             & 20    & 10\\ \Xhline{0.05em}
FlexW   (hard-first)  & 36.71          & 40.92          & 46.32          & 52.54 & 58.62 \\
FlexW   (varied modes) & \textbf{36.76} & \textbf{41.85} & 45.61 & \textbf{52.92} & 57.23\\ \Xhline{0.1em}
\end{tabular}
\end{table}
\begin{table}[t]
\centering
\caption{Accuracies (\%) under the hard-first mode and varied modes of FlexW on CIFAR10. }
\label{tabix10}
\begin{tabular}{lccccc}
\Xhline{0.1em}
Imbalance factor      & 200            & 100            & 50             & 20    & 10\\ \Xhline{0.05em}
FlexW   (hard-first)  & 68.74          & 74.71          & 79.08          & 84.55 & 88.22 \\
FlexW   (varied modes) & \textbf{69.59} & \textbf{75.63} & \textbf{80.43} & 85.03 & 88.00\\ \Xhline{0.1em}
\end{tabular}
\end{table}
\subsection{Standard cifar data sets}
\begin{table*}[b]
\caption{Accuracies (\%) of different methods on WRN-28-2. }
\label{tabelxi11}
\centering
\resizebox{1.9\columnwidth}{!}{
\begin{tabular}{lccccccccccc}
\Xhline{0.1em} 
 Method & Baseline & Focal loss & SPL   & IS    & MentorNet & LOW   & FlexW (easy-first) & FlexW (hard-first) & FlexW (medium-first) & FlexW (both-ends-first)\\ \Xhline{0.05em}  
CIFAR10  & 92.80    & 92.40      & 92.30 & 92.10 & 91.50     & \underline{93.20} & 92.68              & \textbf{94.15}   & 92.87   & 91.34       \\
CIFAR100   & 72.00    & 71.40      & 71.80 & 68.00 & 70.90     & \underline{72.30} & \textbf{72.72}              & 70.22    & 71.45   & 70.14      \\ \Xhline{0.1em} 
\end{tabular}}
\label{tab:plain}
\end{table*}
\begin{table*}[b]
\caption{Accuracies (\%) of different methods on VGG-16.}
\label{tablexii12}
\centering
\resizebox{1.9\columnwidth}{!}{
\begin{tabular}{lcccccccccc}
\Xhline{0.1em}
Data set & Baseline & SPL   & Inverse-SPL & SPLD  & LGL   & Focal loss & FlexW (easy-first) & FlexW (hard-first)& FlexW (medium-first) & FlexW (both-ends-first)\\ 
\Xhline{0.05em}
CIFAR10  & 93.03    & 92.60 & 92.96       & 92.85 & \underline{93.97} & 93.45      & 93.73              & \textbf{94.00}  & 93.28 &92.58\\
CIFAR100 & 71.11    & 70.30 & 70.50       & 70.25 & \underline{74.17} & 74.13      & \textbf{74.96}     & 73.01  & 73.43 & 70.98\\ 
\Xhline{0.1em}
\end{tabular}}
\label{tab:plain}
\end{table*}
\subsubsection{Experiemntal settings}
The standard CIFAR10 and CIFAR100~\cite{AlexKrizhevsky27} data sets are experimented. Experimental settings are the same with those in Section V.A. Focal loss (hard-first)~\cite{TsungYiLin03}, SPL (easy-first)~\cite{MPawanKumar04}, importance sampling~\cite{AngelosKatharopoulos40}, MentorNet~\cite{LuJiang41}, LOW (hard-first)~\cite{CarlosSantiagoa19}, and FlexW (easy-first and hard-first) are compared on WRN-28-2~\cite{SergeyZagoruyko29} as shown in Table~\ref{tabelxi11}. FlexW (easy-first and hard-first) is compared with SPL~\cite{MPawanKumar04} (easy-first), Inverse-SPL~\cite{HaoCheng38} (hard-first), SPLD~\cite{LuJiang39} (easy-first), LGL~\cite{HaoCheng38}, and Focal loss~\cite{TsungYiLin03} (hard-first) on VGG-16~\cite{SimonyanKaren72} in Table~\ref{tablexii12}. 
\subsubsection{Results}
For the standard CIFAR data, there is no prior knowledge or theoretical clues about the learning difficulty distribution which can inspire our choice of the optimal priority mode. Tables~\ref{tabelxi11} and~\ref{tablexii12} show that there does not have a clear judgement among all the priority modes on standard data. Thus, weighting schemes (e.g., FlexW) which can implement four priority modes should be used to achieve different modes and the optimal priority mode can be selected based on the performances on the validation set. 

The experimental results are consistent with the answer for Subproblem (iv). When there is no prior knowledge, theoretical clues, or empirical observations, a weighting function which can achieve four priority modes should be adopted.

\subsection{The selection of hyper-parameters in FlexW}
For ease of use, we give the parameter ranges with stable performances corresponding to the four priority modes which are shown in Table~\ref{tabqujian}. In practical applications, grid search can be used to search parameters within the given parameter ranges. We have verified that the performances of FlexW are stable within these intervals.

\begin{table}[htbp]\scriptsize
\caption{Hyper-parameter intervals in which the performance is stable.}
\label{tabqujian}
\centering
\resizebox{0.98\columnwidth}{!}{
\begin{tabular}{lllll}
\toprule
{Priority mode} & {Easy-first}                                                                                          & {Medium-first}                                                                                        & {Hard-first}     & {Both-ends-first}                                                                                    \\ \midrule

{Intervals}     
& { \begin{tabular}[c]{@{}l@{}}{[}-0.6,-0.2{]}×{[}0.1,0.4{]}\\ {[}0.2,0.6{]}×{[}0.9,1.2{]}\end{tabular}} 
& { \begin{tabular}[c]{@{}l@{}}{[}0.2,0.6{]}×{[}0.4,0.8{]} \end{tabular}}
&{ \begin{tabular}[c]{@{}l@{}}{[}0.2,0.6{]}×{[}0.1,0.4{]}\\ {[}-0.6,-0.2{]}×{[}0.9,1.2{]}\end{tabular}}
& { \begin{tabular}[c]{@{}l@{}}{[}-0.6,-0.2{]}×{[}0.4,0.8{]}
\end{tabular}}\\
\bottomrule                                                       &       
\end{tabular}}
\end{table}
For easy-first and hard-first modes, parameters \{$\gamma = -0.5$, $\alpha = 0.15$\} and \{$\gamma = 0.5$, $\alpha = 0.15$\} can achieve good results in most cases during our experiments. Under the medium-first and two-ends-first modes, the preference for easy or hard samples still exists as shown in Figs.~\ref{fig2}(b) and (d). When $\alpha = 0.58$, the preference for easy and hard samples is approximately equal. Also, the value of $\gamma $ for the medium-first and two-ends-first modes can be set to $0.5$ and $-0.5$. The two-ends-first mode should be used with caution because there are few data sets in the real world that satisfy the corresponding difficulty distribution. 
\section{Answers and discussions}
According to the aforementioned theoretical analyses and empirical observations, a comprehensive answer is obtained for our investigated ``easy-or-hard" question:
\begin{itemize}
\item No universal fixed optimal priority mode exists for an arbitrary learning task. 
\item There are two other typical priority modes, namely, medium-first and two-ends-first.
\item The weight for noisy samples should be reduced. Thus, the priority modes of easy-first and medium-first are more effective on noisy data.
\item The relationship between the difficulty distribution of the training data and priority mode can be analyzed based on Eq.~(3). Four conclusions are obtained:  (1) If the training data has excessive easy samples, hard-first mode is the first choice. (2) If there are excessive hard samples, the easy-first mode should be selected. (3) If the training data has excessive medium-difficult samples, two-ends-first should be adopted. (4) If there are both excessive easy and hard samples, the priority mode of medium-first should be selected. 
\item The priority mode is not necessary to be fixed during training. If the difficulty distribution of the training data changes greatly during the training process, the priority mode should also be changed.
\item If there are not any prior knowledge or theoretical clues, the weighting schemes (e.g., FlexW) which can achieve all the four priority modes should be used. Which mode is appropriate depends on the results of the tuning of the hyper-parameters.

\end{itemize}
The above answer indicates that the measurement of the learning difficulty of samples is crucial as the weight is directly determined by the difficulty distribution of the training set. In most existing studies, the learning difficulty is approximated by the loss (or the predicted probability)~\cite{TsungYiLin03,MPawanKumar04} which is shown in Table~II. Although it is reasonable, an ideal solution should fully consider factors such as loss, the sample’s neighborhood, category distribution, and noise level. This study will be the focus of our future work.

Another important issue is the judgement of whether easy/medium/hard samples are excessive in the training set. Prior knowledge of the training set such as noise rate and category proportion can help us determine this. If there is not any prior knowledge, the excess of easy/medium/hard samples should be judged according to the difference between the distributions of the entire sample space and the training data set. However, it is impractical to utilize the distribution of the entire sample space. A feasible way is to take the distribution of the validation set as a reference 
because in deep learning the validation set is regarded as an unbiased data set.\par
\section{Conclusions}
This study focused on an interesting and important question about the choices of priority modes on easy, medium, and hard samples for learning tasks. A deep investigation for this question facilitates the understanding of various existing weighting schemes and the choice of an appropriate scheme for a new learning task. First, a general optimized objective function is proposed which can mathematically explain the relationship between the difficulty distribution and the priority mode. This general objective function provides a comprehensive view to theoretically analyze the ``easy-or-hard" question. Second, 
several theoretical answers are obtained based on the general objective function. Two other priority modes, namely, medium-first and two-ends-first, are revealed. The priority mode is not necessary to be fixed during training. If the difficulty distribution change greatly, the priority mode should also be changed. 
Third, an effective weighting solution is proposed when there is no prior knowledge and theoretical clues. This solution alleviates the defects of existing methods that only one priority mode can be implemented. 
Fourth, extensive experiments for various tasks under different scenarios 
are conducted under different data characteristics. A comprehensive answer for the ``easy-or-hard" question is obtained according to the theoretical analyses and empirical evaluations.






\section*{Appendix}
\subsection{Other weighting methods}
Apart from weighting schemes investigated in our study, there are also other sample weighting techniques including meta-optimization, teacher-student strategies, and sampling methods.

Meta-optimization leverages an additional unbiased data set to optimize sample weight~\cite{JunShu28,ShuangLi62}. Ren et al.\cite{MengyeRen67} proposed the first meta-optimization method, which assigns weights to training samples on the basis of their gradient directions. Meta-class-weight~\cite{MuhammadAbdullahJamal49} exploits meta-learning to estimate class-wise weights. 
However, meta-optimization methods heavily rely on unbiased data sets which are unavailable in many scenarios. By comparison, 
difficulty-based weighting scheme is easier to implement because an additional data set is not necessary.

The teacher-student strategy uses an additional network as the teacher, with the help of the teacher network’s performance to assign weights to samples in the student network~\cite{YangFan55}. MentorNet~\cite{LuJiang41} uses the teacher network to assign weights to samples in the student network. Samples that are quite hard for the student network will be dropped (weights are set to 0 for these samples) in this case. However, this strategy is computationally expensive and requires an additional network.

Importance sampling aims to reduce the variance of gradient estimates by selecting samples with an adaptive sampling distribution, instead of the traditional uniform sampling~\cite{DeannaNeedell83, PeilinZhao84, Wang90}. However, it needs to know the gradient of loss with respect to each of the network’s parameters and for each training sample before each single gradient descent step. In the context of deep learning, this is computationally infeasible. 
Compared to the sampling-based methods, difficulty-based weighting scheme can process all samples each epoch, thus guaranteeing that we always know how the network is performing on each sample.
\subsection{Theoretical analyses for Propositions 1 and 2}
A strict proof for Proposition 1 is challenging. We give a proof under a special case that the weights exerted on $R_{easy}$ are identical. Without loss of generality, the weights on each sample in $R_{hard}$ are denoted as $(1+\epsilon)$, where $\epsilon >0$.

Let $BiasT(\emph{c})$ and $VarT(\emph{c})$ be the values of bias and variance terms defined in Eq.~(4) in Section~IV-C, respectively, when the model complexity is $\emph{c}$. First, we yield 
\begin{equation}
\begin{aligned}
\frac{\partial Err}{\partial c}\bigg|_{c^{*}} = \frac{\partial BiasT(c)}{\partial c}\bigg|_{c^{*}} + \frac{\partial VarT(c)}{\partial c}\bigg|_{c^{*}}=0
\end{aligned}.
\end{equation}
According to Assumptions 1 and 2, we have
\begin{equation}
\begin{aligned}
\begin{array}{l}
\frac{\partial BiasT_{easy}(c)}{\partial c}\bigg|_{c^{*}} + \frac{\partial VarT_{easy}(c)}{\partial c}\bigg|_{c^{*}}>0\\
\frac{\partial BiasT_{hard}(c)}{\partial c}\bigg|_{c^{*}} + \frac{\partial VarT_{hard}(c)}{\partial c}\bigg|_{c^{*}}<0
\end{array}
\end{aligned}.
\end{equation}
Let $p_{easy}$, $p_{medium}$, $p_{hard}$ be the proportions of samples in $R_{easy}$, $R_{medium}$, $R_{hard}$, respectively. We yield
\begin{small}
\begin{equation}
\begin{aligned}
BiasT(c^*) &= p_{easy}BiasT_{easy}(c^*) + p_{medium}BiasT_{medium}{(c^*)} \\
&+ p_{hard}BiasT_{hard}(c^*)\\
VarT(c^*) &= p_{easy}VarT_{easy}(c^*) + p_{medium}VarT_{medium}(c^*)\\ &+ p_{hard}VarT_{hard}(c^*)
\end{aligned}.
\end{equation}
\end{small}When the weights $(1+\epsilon)$ are exerted on $R_{hard}$, then $BiasT(c^*)$ and $VarT(c^*)$ become
\begin{small}
\begin{equation}
\begin{aligned}
BiasT_\epsilon(c^*) &=p_{easy}BiasT_{easy}(c^*) + p_{medium}BiasT_{medium}(c^*)\\& + p_{hard}BiasT_{hard} +  \epsilon p_{hard}BiasT_{hard}(c^*)\\
VarT_\epsilon (c^*) &=p_{easy}VarT_{easy}(c^*) + p_{medium}VarT_{medium}(c^*)
\\  &+ p_{hard}VarT_{hard}(c^*) + \epsilon p_{hard}VarT_{hard}(c^*)
\end{aligned}.
\end{equation}
\end{small}
Based on Eqs.~(13) and (15), we have 
\begin{equation}
\begin{aligned}
\frac{\partial BiasT_{\epsilon}(c)}{\partial c}\bigg|_{c^{*}} + \frac{\partial VarT_{\epsilon}(c)}{\partial c}\bigg|_{c^{*}}<0
\end{aligned}.
\end{equation}
Accordingly, to attain the new balance between the bias and variance terms, the model complexity should be increased. Alternatively, the new optimal model complexity $c^{*}_{new}$ will be larger than $c^{*}$.

The proof for Proposition 2 is with a similar inference manner.

\subsection{Gradient analysis of FlexW}
The weighting function affects the training process by influencing the gradients of samples. To better understand FlexW, the loss gradient of loss function with FlexW is analyzed. The loss gradient is:
\begin{equation}
\begin{aligned}
\frac{{d{\mathcal{L}}}}{{dz}} &= \frac{\partial {\mathcal{L}}}{{\partial p}} \times \frac{{\partial p}}{{\partial z}} \\&= p(1 - p){(1 - p + \alpha )^{\gamma - 1}}{e^{ - \gamma(1 - p + \alpha )}} \times \\&(\gamma\log p(p - \alpha) - \frac{{1 - p + \alpha }}{p})
\end{aligned},
\end{equation}
where $p = \frac{1}{{1 + {e^{ - z}}}}$. The loss gradient of loss function with FlexW is in comparison to the gradients of cross entropy (CE) loss, Focal loss~\cite{TsungYiLin03}, loss function with SPL\_Log~\cite{LuJiang06}, and ASL~\cite{EmanuelBenBaruch10}.
\begin{figure*}[t]
    \centering
    \includegraphics[width=1\linewidth]{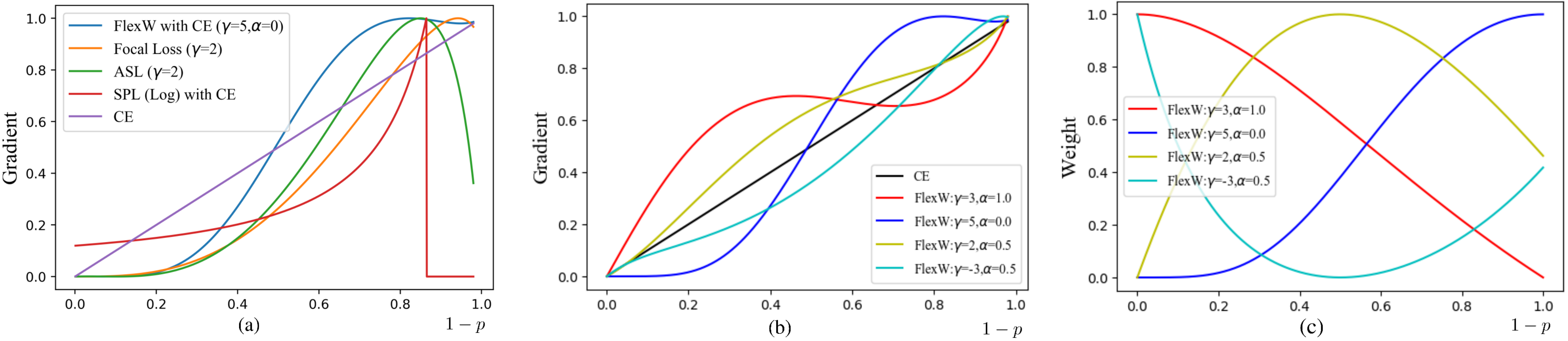} 
   \vspace{-0.22in} 
    \caption{Gradients of different losses.
}
\label{tidu}
\end{figure*}
Fig.~\ref{tidu}(a) shows the gradients of different losses. Under CE loss, harder samples have larger gradients than easier ones. Focal loss increases the gradients of hard samples. However, it is sensitive to noise. ASL decreases the gradients of quiet-hard samples. Fig.~\ref{tidu}(b) shows the gradients of the four variants of FlexW with CE loss. The weight curves of the four variants are shown in Fig.~\ref{tidu}(c). 
When the easy/medium/hard/two-ends-first mode of FlexW is used, the loss gradients of easy/medium/hard/both easy and hard samples are increased compared with those under CE loss shown in Fig.~\ref{tidu}(b). 

\subsection{Node classification for graph data sets}
The sample's neighbors influence the learning difficulty of the sample. This experiment verifies the performances of the easy/hard-first modes on graph data.
\subsubsection{Experimental settings}
Five benchmark graph data sets are used, namely, Cora, Citeseer, Pubmed, Coauthor CS, and Coauthor Physics~\cite{ZhilinYang32,OleksandrShchur33}. The basic model is an eight-layer GCN~\cite{JoanBruna34}. Transductive training is used and all node features are accessible during training. We apply the full-supervised training setting used in ~\cite{WenbingHuang68} and ~\cite{JieChen69} on all data sets. During training, Adam~\cite{DiederikPKingma70} is used as the optimizer. The value of the learning rate is $0.001$. The weight decay is set to $5 \times 10^{-4}$. The value of the epoch is set to $400$. The dimension of the hidden layers is set to $128$. The settings for parameters in FlexW are the same as Section~V-A.

\subsubsection{Results}
\begin{table}[htbp]\scriptsize
\caption{Accuracies (\%) of the competing methods on five graph data sets.}
\label{tab8}
\centering
\resizebox{0.98\columnwidth}{!}{
\begin{tabular}{lccccccccc}
\Xhline{0.1em}  
Method\textbackslash{}Data   set & Cora  & Citeseer & Pubmed & Coauthor CS & Coauthor Physics \\ \Xhline{0.05em}  
Original                         & 86.50 & 78.70    & \underline{90.90}  & 90.70       & 94.00  \\
SPL\_Poly                        & \underline{87.10} & 78.30    & 90.40  & 92.07       & \underline{95.78} \\
SPL\_Log                         & \underline{87.10} & 78.30    & 90.20  & \underline{93.44}       & 95.65 \\
SPL\_Binary                      & 86.50 & \underline{78.90}    & 89.90  & 93.16       & 94.48 \\
Focal loss                       & 86.10 & 78.70    & 89.70  & 89.43       & 93.03 \\
FlexW (hard-first)               & 86.60 & 78.10    & 90.00  & 89.34       & 93.90 \\
FlexW (easy-first)               & \textbf{87.50} & \textbf{79.50}    & \textbf{91.30}  & \textbf{93.71}       & \textbf{95.85} \\
\Xhline{0.1em} 
\end{tabular}}
\end{table}
In GCN, the heterogeneous nodes around a node negatively affect the representation of that node. To study the performances of the easy-first and hard-first modes on graph data, FlexW (both easy/hard-first) is compared with several variations of SPL and Focal loss.

\begin{figure}[t]
    \centering
    \includegraphics[width=0.7\linewidth]{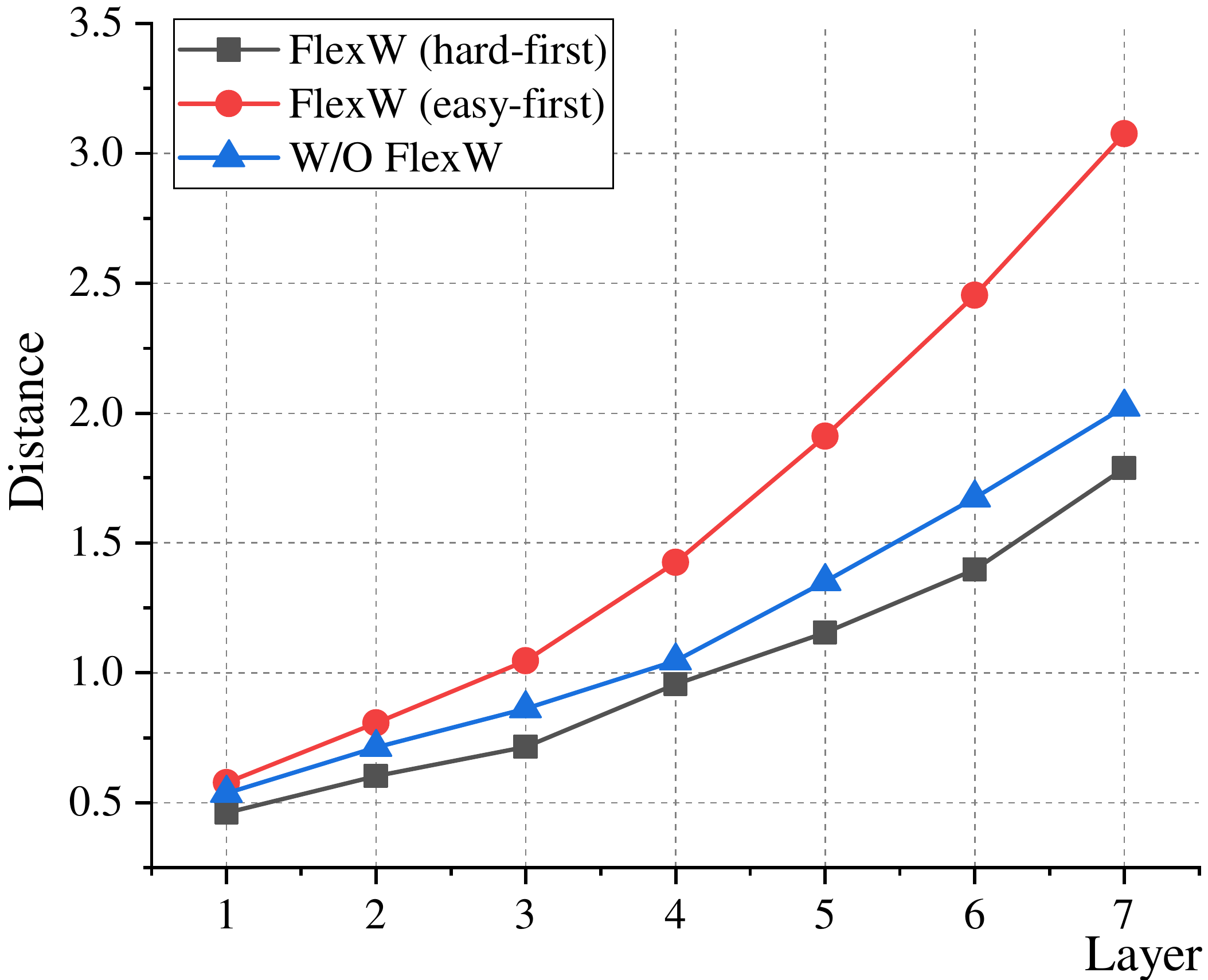} 
    \vspace{-0.12in} 
    \caption{The degree of over-smoothing under different priority modes.
}
\label{over}
\end{figure}

Table~\ref{tab8} shows the results of the competing methods. In general, the easy-first schemes (i.e., SPL and FlexW (easy-first)) perform better than the hard-first ones (i.e., Focal loss and FlexW (hard-first)). As the hard samples in a graph are mostly those with a large proportion of heterogeneous adjacent nodes, easy-first schemes can reduce the negative influence of the information exchange among heterogeneous nodes. To investigate it, the over-smoothing degree is measured by computing the Euclidean distance between the output of the current layer and that of the previous one~\cite{YuRong76}. The smaller the distance is, the more serious the over-smoothing is~\cite{DeliChen35}. Fig.~\ref{over} indicates that the distances of the model with FlexW (easy-first) are larger than those of the baseline and hard-first. Thus, the easy-first mode can relieve the over-smoothing phenomenon while the hard-first mode will exacerbate this phenomenon.

\vfill


\begin{thebibliography}{1}
\bibliographystyle{IEEEtran}
\bibitem{fraud001}
A. Dal Pozzolo, G. Boracchi, O. Caelen, C. Alippi and G. Bontempi,``Credit Card Fraud Detection: A Realistic Modeling and a Novel Learning Strategy," in \textit{IEEE Transactions on Neural Networks and Learning Systems}, vol. 29, no. 8, pp. 3784-3797, Aug. 2018, doi: 10.1109/TNNLS.2017.2736643.
\bibitem{medical}
Artan, Yusuf and Haider, Masoom A. and Langer, Deanna L. and van der Kwast, Theodorus H. and Evans, Andrew J. and Yang, Yongyi and Wernick, Miles N. and Trachtenberg, John and Yetik, Imam Samil.,``Prostate Cancer Localization With Multispectral MRI Using Cost-Sensitive Support Vector Machines and Conditional Random Fields," in \textit{IEEE Transactions on Image Processing}, vol. 19, no. 9, pp. 2444-2455, Sept. 2010.
\bibitem{BuyuLi11}
B. Li, Y. Liu, and X. Wang. ``Gradient Harmonized Single-stage Detector,'' in \textit{33rd AAAI Conference on Artificial Intelligence}, pp. 8577--8584, 2019.
\bibitem{Verleysen91}
Frénay B, Verleysen M. ``Classification in the presence of label noise: a survey." \textit{IEEE Transactions on Neural Networks and Learning Systems}, vol. 25, No. 5, pp. 845-869, 2013.
\bibitem{DeliChen35}
D. Chen, Y. Lin, W. Li, P. Li, J. Zhou, and X. Sun. ``Measuring and relieving the over-smoothing problem for graph neural networks from the topological view,'' in \textit{Proceedings of the 34th AAAI Conference on Artificial Intelligence}, pp. 3438--3445, 2020.
\bibitem{JingfengZhang16}
J. Zhang, J. Zhu, G. Niu, B. Han, M. Sugiyama, and M. Kankanhalli. ``Geometry-aware Instance-reweighted Adversarial Training,'' in \textit{Proceedings of the 9th International Conference on Learning Representations}, Online, pp. 1--29, 2021.
\bibitem{detectionsurvey}
Z. Zhao, P. Zheng, S. Xu and X. Wu, "Object Detection With Deep Learning: A Review," in \textit{IEEE Transactions on Neural Networks and Learning Systems}, vol. 30, no. 11, pp. 3212-3232, Nov. 2019.
\bibitem{YinCui17}
Y. Cui, M. Jia, T. Lin, Y. Song, and S. Belongie. ``Class-Balanced Loss Based on Effective Number of Samples,'' in \textit{2019 IEEE Conference on Computer Vision and Pattern Recognition}, pp. 9260--9269, 2019.
\bibitem{TsungYiLin03}
T. Lin, P. Goyal, R. Girshick, K. He, and P. Dollar. ``Focal Loss for Dense Object Detection,'' in \textit{2017 IEEE International Conference on Computer Vision}, Venice, Italy, pp. 2999--3007, 2017.
\bibitem{MPawanKumar04}
M. Pawan Kumar, B. Packer, and D. Koller. ``Self-paced learning for latent variable models,'' in \textit{24th Annual Conference on Neural Information Processing Systems}, Red Hook, New York, USA, pp. 1--9, 2010.
\bibitem{XinWang22}
X. Wang, Y. Chen, and W. Zhu. ``A Survey on Curriculum Learning,'' \textit{IEEE Transactions on Pattern Analysis and Machine Intelligence}, vol. 1, no. 1, pp. 33788677, 2021.
\bibitem{LuJiang06}
L. Jiang, D. Meng, T. Mitamural, and A. G. Hauptmann. ``Easy samples first: Self-paced reranking for zero-example multimedia search,'' in \textit{2014 ACM Conference on Multimedia}, pp. 547--556, 2014.
\bibitem{MaciejZieba07}
M. Zieba, J. M. Tomczak, and J. swiatek. ``Self-paced Learning for Imbalanced Data," in \textit{Asian Conference on Intelligent Information and Database Systems}, pp. 564--573, 2016.
\bibitem{XiangLi09}
X. Li, W. Wang, L. Wu, S. Chen, X. Hu, J. Li, J. Tang, and J. Yang. ``Generalized Focal Loss: Learning Qualified and Distributed Bounding Boxes for Dense Object Detection,'' \textit{arXiv preprint arXiv:2006.04388}, pp. 1--14, 2020.
\bibitem{EmanuelBenBaruch10}
E. Ben-Baruch, T. Ridnik, N. Zamir, A. Noy, I. Friedman, M. Protter, and L. Zelnik-Manor. ``Asymmetric Loss For Multi-Label Classification,'' \textit{arXiv preprint arXiv:2009.14119}, pp. 1--12, 2020.
\bibitem{YoavFreund12}
Y. Freund, and R. E. Schapire. ``Experiments with a New Boosting Algorithm,'' in \textit{Machine Learning: Proceedings of the Thirteenth International Conference}, pp. 1--9, 1996.
\bibitem{SongyangZhang14}
S. Zhang, Z. Li, S. Yan, X. He, and J. Sun. ``Distribution Alignment: A Unified Framework for Long-tail Visual Recognition,'' \textit{arXiv preprint arXiv:2103.16370}, pp. 1--10, 2021.
\bibitem{WenjieWang18}
W. Wang, F. Feng, X. He, L. Nie, and T. Chua. ``Denoising Implicit Feedback for Recommendation,'' in \textit{Proceedings of the 14th ACM International Conference on Web Search and Data Mining}, pp. 373--381, 2021.
\bibitem{CarlosSantiagoa19}
C. Santiagoa, C. Barataa,  M. Sasdellib, G. Carneirob, and J. C.Nasciment. ``LOW: Training deep neural networks by learning optimal sample weights,'' \textit{Pattern Recognition}, vol. 110, no. 1, pp. 1--12, 2021.
\bibitem{EvanZheranLiu20}
E. Z. Liu, B. Haghgoo, A. S. Chen, A. Raghunathan, P. W. Koh, S. Sagawa, P. Liang, and C. Finn. ``Just Train Twice: Improving Group Robustness without Training Group Information,'' \textit{arXiv preprint arXiv:2107.09044}, pp. 1--16, 2021.
\bibitem{ThibaultCastells21}
T. Castells, P. Weinzaepfel, and J. Revaud. ``SuperLoss: A generic loss for robust curriculum learning,'' in \textit{Proceedings of the 34th Conference on Neural Information Processing Systems}, pp. 1--12, 2020.
\bibitem{Fernando_imb}
K. R. M. Fernando and C. P. Tsokos, ``Dynamically Weighted Balanced Loss: Class Imbalanced Learning and Confidence Calibration of Deep Neural Networks," in \textit{IEEE Transactions on Neural Networks and Learning Systems}, pp. 2162-2388, 2021.
\bibitem{YoshuaBengio15}
Y. Bengio, J. Louradour, R. Collobert, J. Weston. ``Curriculum learning,'' in \textit{Proceedings of the 26th International Conference on Machine Learning}, pp. 41--48, 2009.
\bibitem{WonyoungShin79}
W. Shin, J. Ha, S. Li, Y. Cho, H. Song, and S. Kwon. ``Which Strategies Matter for Noisy Label Classification? Insight into Loss and Uncertainty,'' \textit{arXiv preprint arXiv:2008.06218}, pp. 1--14, 2020.
\bibitem{SalmanHKhan44}
S. H. Khan, M. Hayat, M. Bennamoun, F. Sohel, and R. Togneri. ``Cost-sensitive learning of deep feature representations from imbalanced data,'' \textit{IEEE Transactions on Neural Networks and Learning Systems}, vol. 29, no. 8, pp. 3573--3587, 2018.
\bibitem{TimvanErven88}
T. V. Erven, and P. Harremos. ``Renyi Divergence and Kullback-Leibler Divergence,'' \textit{IEEE Transactions on Information Theory}, vol. 60, no. 7, pp. 3797--3820, 2014.
\bibitem{JufengYang23}
J. Yang, X. Wu, J. Liang, X. Sun, M. Cheng, P. L. Rosin, and L. Wang., ``Self-Paced Balance Learning for Clinical Skin Disease Recognition," in IEEE Transactions on Neural Networks and Learning Systems, vol. 31, no. 8, pp. 2832-2846, Aug. 2020, doi: 10.1109/TNNLS.2019.2917524.
\bibitem{ZitongYang82}
Z. Yang, Y. Yu, C. You, J. Steinhardt, and Y. Ma. ``Rethinking Bias-Variance Trade-off for Generalization of Neural Networks,'' in \textit{Proceedings of the 37the International Conference on Machine Learning}, pp. 10698--10708, 2020.
\bibitem{PedroDomingos25}
P. Domingos. ``A Unified Bias-Variance Decomposition for Zero-One and Squared Loss,'' in \textit{Proceedings of the Seventeenth National Conference on Artificial Intelligence and Twelfth Conference on Innovative Applications of Artificial Intelligence}, pp. 564--569, 2000.
\bibitem{ScottFortmannRoe81}
S. Fortmann-Roe. ``Understanding the Bias-Variance Tradeoff,'' \textit{\url{http://scott.fortmann-roe.com/docs/BiasVariance.html}}, 2012.
\bibitem{XiaoxiaWu31}
X. Wu, E. Dye,r and B. Neyshabur. ``When Do Curricula Work?,'' in \textit{Proceedings of the 9th International Conference on Learning Representations}, pp. 1--23, 2021.
\bibitem{GuyHacohen88}
G. Hacohen, and D. Weinshall. ``On The Power of Curriculum Learning in Training Deep Networks,'' \textit{arXiv preprint arXiv:1904.03626 }, pp. 1--13, 2019.
\bibitem{AlexKrizhevsky27}
A. Krizhevsky, {\it{Learning multiple layers of features from tiny images}}, MIT Press. Massachusetts, 2009.
\bibitem{JunShu28}
J. Shu, Q. Xie, L. Yi, Q. Zhao, S. Zhou, Z. Xu, and D. Meng. ``Meta-weight-net: Learning an explicit mapping for sample weighting,'' in \textit{Proceedings of the 33rd Annual Conference on Neural Information Processing Systems}, pp. 1--23, 2019.
\bibitem{SergeyZagoruyko29}
S. Zagoruyko, and N. Komodakis. ``Wide Residual Networks,'' in \textit{Proceedings of the British Machine Vision Conference}, pp. 87.1--87.12, 2016.
\bibitem{KaimingHe30}
K. He, X. Zhang, S. Ren, and J. Sun. ``Deep Residual Learning for Image Recognition,'' in \textit{2016 IEEE Conference on Computer Vision and Pattern Recognition}, pp. 770--778, 2016.
\bibitem{ScottReed60}
S. Reed, H. Lee, D. Anguelov, C. Szegedy, D. Erhana, and A. Rabinovich. ``Training Deep Neural Networks on Noisy Labels with Bootstrapping,'' in \textit{Proceedings of the 3rd International Conference on Learning Representations}, pp. 1--11, 2015.
\bibitem{JacobGoldberger61}
J. Goldberger, and E. Ben-Reuven. ``Training deep neural-networks using a noise adaptation layer,'' in \textit{Proceedings of the 5th International Conference on Learning Representations}, pp. 1--9, 2017.
\bibitem{BoHan58}
B. Han, Q. Yao, X. Yu, G. Niu, M. Xu, W. Hu,  I. W. Tsang, and M. Sugiyama. ``Co-teaching: Robust Training of Deep Neural Networks with Extremely Noisy Labels,'' in \textit{Proceedings of the 32nd Conference on Neural Information Processing Systems}, pp. 8536--8546, 2018.
\bibitem{XingjunMa59}
X. Ma, Y. Wang, M. E. Houle, S. Zhou, S. Erfani, S. Xia, S. Wijewickrema, and J. Bailey. ``Dimensionality-Driven Learning with Noisy Labels,'' in \textit{Proceedings of the 35th International Conference on Machine Learning}, pp. 3361--3370, 2018.
\bibitem{LuJiang41}
L. Jiang, Z. Zhou, T. Leung, L. Li, and F. Li. ``Mentornet: Learning data-driven curriculum for very deep neural networks on corrupted labels,'' in \textit{Proceedings of the 35th International Conference on Machine Learning}, pp. 3601--3620, 2018.
\bibitem{HongyiZhang63}
H. Zhang, M. Cisse, Y. N. Dauphin, and D. Lopez-Paz. ``Mixup: BEYOND EMPIRICAL RISK MINIMIZATION,'' in \textit{Proceedings of the 6th International Conference on Learning Representations}, pp. 1--13, 2018.
\bibitem{KaidiCao64}
K. Cao, C. Wei, A. Gaidon, N. Arechiga, and T. Ma. ``Learning Imbalanced Datasets with Label-Distribution-Aware Margin Loss,'' in \textit{Proceedings of the 33rd International Conference on Neural Information Processing Systems}, pp. 1567--1578, 2019.
\bibitem{JingruTan65}
J. Tan, C. Wang, B. Li, Q. Li, W. Ouyang, C. Yin, and J. Yan. ``Equalization Loss for Long-Tailed Object Recognition,'' in \textit{2020 IEEE/CVF Conference on Computer Vision and Pattern Recognition}, pp. 11659--11668, 2020.
\bibitem{MengyeRen67}
M. Ren and W. Zeng and B. Yang and R. Urtasun. ``Learning to reweight examples for robust deep learning,'' in \textit{Proceedings of the 35th International Conference on Machine Learning}, pp. 6900--6909, 2018.
\bibitem{YinCui66}
Y. Cui, Y. Song, C. Sun, A. Howard, and S. Belongie. ``Large Scale Fine-Grained Categorization and Domain-Specific Transfer Learning,'' in \textit{2018 IEEE/CVF Conference on Computer Vision and Pattern Recognition}, pp. 4109--4118, 2018.
\bibitem{EveringhamMark36}
E. Mark,  G. Luc, C. K. I. Williams, W. John, and Z. Andrew. ``The PASCAL Visual Object Classes (VOC) Challenge,'' \textit{International Journal of Computer Vision}, vol. 5, no. 1, pp. 329--359, 1996.
\bibitem{EveringhamMark37}
E. Mark, E. S. M. AliVan,  G. Luc,  C. K. I.Williams, W. John, and Z. Andrew. ``The PASCAL Visual Object Classes Challenge: A Retrospective,'' \textit{International Journal of Computer Vision}, vol. 111, no. 1, pp. 98--136, 2015.
\bibitem{AlexeyBochkovskiy75}
A. Bochkovskiy, C. Wang, and H. M. Liao. ``YOLOv4: Optimal Speed and Accuracy of Object Detection,'' \textit{arXiv preprint arXiv:2004.10934}, pp. 1--17, 2020.
\bibitem{JosephRedmon89}
J. Redmon, S. Divvala, R. Girshick, and A. Farhadi. `You Only Look Once: Unified, Real-Time Object Detection,'' in \textit{2016 IEEE Conference on Computer Vision and Pattern Recognition}, Las Vegas, Nevada, USA, pp. 779--788, 2016.
\bibitem{AngelosKatharopoulos40}
A. Katharopoulos, and F. Fleuret. ``Not All Samples Are Created Equal: Deep Learning with Importance Sampling,'' in \textit{Proceedings of the 35th International Conference on Machine Learning}, pp. 1--13, 2018.
\bibitem{HaoCheng38}
H. Cheng, D. Lian, B. Deng, S. Gao, T. Tan, and Y. Geng. ``Local to global learning: Gradually adding classes for training deep neural networks,'' in \textit{Proceedings of the 32nd IEEE/CVF Conference on Computer Vision and Pattern Recognition}, pp. 4743--4751, 2019.
\bibitem{LuJiang39}
L. Jiang, D. Meng, S. Yu, Z. Lan, S. Shan, and A. G. Hauptmann. ``Self-Paced Learning with Diversity,'' in \textit{Proceedings of the 28th Annual Conference on Neural Information Processing Systems}, pp. 2078--2086, 2014.
\bibitem{SimonyanKaren72}
S. Karen, and Z. Andrew. ``Very Deep Convolutional Networks for Large-Scale Image Recognition,'' \textit{arXiv preprint arXiv:1409.1556}, pp. 1--10, 2014.
\bibitem{ShuangLi62}
S. Li,  K. Gong, C. H. Liu, Y. Wang, F. Qiao, and X. Cheng. ``MetaSAug: Meta Semantic Augmentation for Long-Tailed Visual Recognition,'' \textit{arXiv preprint arXiv:2103.12579}, pp. 1--10, 2021.
\bibitem{MuhammadAbdullahJamal49}
M. A. Jamal, M. Brown, M. Yang, L. Wang, and B. Gong. ``Rethinking Class-Balanced Methods for Long-Tailed Visual Recognition from a Domain Adaptation Perspective,'' in \textit{2020 IEEE/CVF Conference on Computer Vision and Pattern Recognition}, pp. 7607--7616, 2020.
\bibitem{YangFan55}
Y. Fan, F. Tian, T. Qin, X. Li, and T. Liu. ``Learning to teach,'' in \textit{Proceedings of the 6th International Conference on Learning Representations}, San Diego, California, USA, pp. 1--16, 2018.
\bibitem{Wang90}
Wang Z, Li H X, Chen C. ``Incremental reinforcement learning in continuous spaces via policy relaxation and importance weighting." \textit{IEEE Transactions on Neural Networks and Learning Systems}, vol. 31, No. 6, pp. 1870-1883, 2019.
\bibitem{DeannaNeedell83}
D. Needell, N. Srebro, and R. Ward. ``Stochastic gradient descent, weighted sampling, and the randomized Kaczmarz algorith,'' in \textit{Proceedings of the 28th Annual Conference on Neural Information Processing Systems}, pp. 1017--1025, 2014.
\bibitem{PeilinZhao84}
P. Zhao, and T. Zhang. ``Stochastic Optimization with Importance Sampling,'' in \textit{Proceedings of the 32th International Conference on Machine Learning}, pp. 1--9, 2015.
\bibitem{ZhilinYang32}
Z. Yang, W. W. Cohen, and R. Salakhutdinov. ``Revisiting semi-supervised learning with graph embeddings,'' in \textit{Proceedings of the 33th International Conference on Machine Learning}, pp. 40--48, 2016.
\bibitem{OleksandrShchur33}
O. Shchur, M. Mumme, A. Bojchevski, and S. Günnemann. ``Pitfalls of Graph Neural Network Evaluation,'' in \textit{Proceedings of the 32nd Conference on Neural Information Processing Systems}, pp. 1--11, 2018.
\bibitem{JoanBruna34}
J. Bruna, W. Zaremba, A. Szlam, and Y. LeCun. ``Spectral networks and deep locally connected networks on graphs,'' in \textit{Proceedings of the 2nd International Conference on Learning Representations}, pp. 1--14, 2014.
\bibitem{WenbingHuang68}
W. Huang, T. Zhang, Y. Rong, and  J. Huang. ``Adaptive sampling towards fast graph representation learning,'' in \textit{Proceedings of the 32nd Conference on Neural Information Processing Systems}, pp. 4558--4567, 2018.
\bibitem{JieChen69}
J. Chen, T. Ma, and C. Xiao. ``FastGCN: Fast learning with graph convolutional networks via importance sampling,'' in \textit{Proceedings of the 6th International Conference on Learning Representations}, pp. 1--15, 2018.
\bibitem{DiederikPKingma70}
D. P. Kingma, and J. L. Ba. ``Adam: A method for stochastic optimization,'' in \textit{Proceedings of the 3rd International Conference on Learning Representations}, pp. 1--15, 2015.
\bibitem{YuRong76}
Y. Rong, W. Huang, T. Xu, and J. Huang. ``DropEdge: Towards Deep Graph Convolutional Networks on Node Classification,'' in \textit{Proceedings of the 8th International Conference on Learning Representations}, pp. 2261--2269, 2017.















































































































%



\end{thebibliography}
\end{document}